\documentclass[10pt,journal,compsoc]{IEEEtran}

\usepackage{etex}

\usepackage{amsthm}
\usepackage{amssymb}
\usepackage{graphicx}
\usepackage{subfigure}
\usepackage{url}
\usepackage[ruled,vlined,boxed,linesnumbered]{algorithm2e}
\usepackage[usenames,dvipsnames]{xcolor}

\usepackage[cmex10]{amsmath}\interdisplaylinepenalty=2500

\usepackage{tabulary}
\usepackage{verbatim}
\usepackage{multirow}
\usepackage{wrapfig}
\usepackage{paralist}
\usepackage{listings}
\usepackage{mathptmx}
\usepackage{cite}
\usepackage{booktabs}
\usepackage{caption}

\usepackage{tikz}
\usetikzlibrary{datavisualization}
\usetikzlibrary{datavisualization.formats.functions}
\usetikzlibrary{positioning, calc, backgrounds, arrows}
\usetikzlibrary{shapes.geometric}
\usetikzlibrary{decorations.pathreplacing}
\usetikzlibrary{patterns}

\usepackage{pgfplots}

\usepackage{filecontents}

\makeatletter
\pgfdeclaregenericanchor{r}{
  \pgf@sh@reanchor{\csname pgf@sh@ns@\pgfreferencednodename\endcsname}{center}
  {
    \pgfsettransform{\csname pgf@sh@nt@\pgfreferencednodename\endcsname}
    \pgf@pos@transform{\pgf@x}{\pgf@y}
    \global\pgf@x=\pgf@x
    \global\pgf@y=\pgf@y
  }
  {
    \pgf@pos@transform{\pgf@x}{\pgf@y}
    \global\pgf@x=\pgf@x
    \global\pgf@y=\pgf@y
  }
  {
    \pgfsettransform{\csname pgf@sh@nt@\pgfreferencednodename\endcsname}
    \pgftransforminvert
    \pgf@pos@transform{\pgf@x}{\pgf@y}
    \global\pgf@x=\pgf@x
    \global\pgf@y=\pgf@y
  }
}
\makeatother

\definecolor{tmpRed}{RGB}{215,25,28}
\definecolor{tmpOrange}{RGB}{253,174,97}
\definecolor{tmpGreen}{RGB}{171,221,164}
\definecolor{tmpBlue}{RGB}{43,131,186}
    
\newlength{\starsize}
\newlength{\starspread}
\tikzset{starsize/.code={\setlength{\starsize}{#1}},
         starspread/.code={\setlength{\starspread}{#1}}}
\tikzset{starsize=1mm,
         starspread=3mm}
\pgfdeclarepatternformonly[\starspread,\starsize]
  {custom fivepointed stars}
  {\pgfpointorigin}
  {\pgfqpoint{\starspread}{\starspread}}
  {\pgfqpoint{\starspread}{\starspread}}
  {
   \pgftransformshift{\pgfqpoint{\starsize}{\starsize}}
   \pgfpathmoveto{\pgfqpointpolar{18}{\starsize}}
   \pgfpathlineto{\pgfqpointpolar{162}{\starsize}}
   \pgfpathlineto{\pgfqpointpolar{306}{\starsize}}
   \pgfpathlineto{\pgfqpointpolar{90}{\starsize}}
   \pgfpathlineto{\pgfqpointpolar{234}{\starsize}}
   \pgfpathclose%
   \pgfusepath{fill}
}

\pgfdeclarepatternformonly{custom horizontal lines}{\pgfpointorigin}{\pgfqpoint{50pt}{1pt}}{\pgfqpoint{50pt}{3pt}}%
{
  \pgfsetlinewidth{0.2pt}
  \pgfpathmoveto{\pgfqpoint{0pt}{0.5pt}}
  \pgfpathlineto{\pgfqpoint{100pt}{0.5pt}}
  \pgfusepath{stroke}
}

\pgfdeclarepatternformonly{custom north east lines}{\pgfqpoint{0pt}{0pt}}{\pgfqpoint{4pt}{4pt}}{\pgfqpoint{3.6pt}{3.6pt}}%
{
  \pgfsetlinewidth{0.2pt}
  \pgfpathmoveto{\pgfqpoint{0pt}{0pt}}
  \pgfpathlineto{\pgfqpoint{6.2pt}{6.2pt}}
  \pgfusepath{stroke}
}

\pgfdeclarepatternformonly{custom north west lines}{\pgfqpoint{0pt}{0pt}}{\pgfqpoint{4pt}{4pt}}{\pgfqpoint{3.6pt}{3.6pt}}%
{
  \pgfsetlinewidth{0.2pt}
  \pgfpathmoveto{\pgfqpoint{0pt}{0pt}}
  \pgfpathlineto{\pgfqpoint{6.2pt}{-6.2pt}}
  \pgfusepath{stroke}
}

\newtheorem{definition}{Definition}

\graphicspath{{images/}}
\begin{document}

\title{Detecting Sudden and Gradual Drifts in Business Processes from Execution Traces
}


\author{
  Abderrahmane Maaradji,
  Marlon Dumas,
  Marcello La Rosa and 
  Alireza Ostovar
    \IEEEcompsocitemizethanks{
        \IEEEcompsocthanksitem A. Maaradji, M. La Rosa and A. Ostovar are with the Queensland University of Technology, Australia
        \protect\\ Email: \{abderrahmane.maaradji,m.larosa,alireza.ostovar\}@qut.edu.au            \IEEEcompsocthanksitem M. Dumas is with the University of Tartu, Estonia.
        \protect\\ Email: marlon.dumas@ut.ee           
    }
}

\IEEEtitleabstractindextext{
\begin{abstract} 
Business processes are prone to unexpected changes, as process workers may suddenly or gradually start executing a process differently in order to adjust to changes in workload, season or other external factors.
Early detection of business process changes enables managers to identify and act upon changes that may otherwise affect process performance. 
Business process drift detection refers to a family of methods to detect changes in a business process by analyzing event logs 
extracted from the systems that support the execution of the process. 
Existing methods for business process drift detection are based on an explorative analysis of a potentially large feature space and in some cases they require users to manually identify specific features that characterize the drift. Depending on the explored feature space, these methods miss various types of changes. Moreover, they are either designed to detect sudden drifts or gradual drifts but not both. This paper proposes an automated and statistically grounded method for detecting sudden and gradual business process drifts under a unified framework. 
An empirical evaluation shows that the method detects typical change patterns with significantly higher accuracy and lower detection delay than existing methods, while accurately distinguishing between sudden and gradual drifts.
\end{abstract}

}

\maketitle

\sloppy
\section{Introduction}
\label{sec:introduction}


Business processes need to continuously change in response to external factors, such as variations in demand, supply, customer expectations and regulations, as well as seasonal factors.
Some changes are planned and documented, but others occur unexpectedly and may remain unnoticed. The latter may be the case for example of changes initiated by individual workers in order to adapt to variations in workload, changes brought about by replacement of human resources, changes in the frequency of certain types of (problematic) cases, or exceptions that in some cases give rise to new workarounds that over time solidify into norms. Such undocumented changes may over time have a significant impact on the performance of the business process.

In this setting, business process managers require methods and tools that allow them to detect and pinpoint changes as early as possible. \emph{Business process drift detection}~\cite{bose2011handling,carmona2012online,bose2014dealing} is a family of techniques to analyze event logs or event streams generated during the execution of a business process in order to detect points in time when the behavior of recent executions of the process (a.k.a. \emph{cases}) differs significantly from that of older one.
The input of such methods is a set of traces, each representing the sequence of events generated by one case of the business process. 

Given this input, existing methods for business process drift detection extract patterns (a.k.a.\ \emph{features}) characterizing each trace. One possible feature is for example that task $A$ occurs before task $B$ in the trace, or that $B$ occurs more than once in the trace. To achieve a suitable level of accuracy, existing techniques either explore large feature spaces automatically or they require the users themselves to identify a specific set of features that are likely to characterize the drift -- implying that the user already has an \emph{a priori} idea of the characteristics of drift. In all cases, these methods will miss a change if it is not characterized by the features employed during the analysis. Furthermore, the scalability of these techniques is hindered by the need to extract and analyze a large set of high-dimensional feature vectors. 

Another limitation of existing methods is that they are designed to detect either gradual drifts or sudden drifts, but not both. 
In this context, a gradual drift is one that does not affect all cases at once. Instead, it initially affects some cases but not others and then gradually affects more cases until it becomes visible across the board.  For example, a new policy in an insurance company may require claim handlers to perform a new check on each insurance claim. The insurance company may decide to instantly apply the check to all claims (sudden drift), or may first start by performing the check on short-term and low-value claims and over time extend it to long-term and high-value claims (gradual drift).

Discerning gradual drifts is more challenging than sudden drifts as there is no time point such that cases completed after this point totally differ from cases completed before it. In contrast to a sudden drift where an abrupt change occurs at a time point, a gradual drift is characterized by a transition time interval where the prior process behavior fades-out whereas the posterior one fades-in. While a specialized technique for gradual drift detection has been proposed in~\cite{Martjushev2015}, this latter is not integrated with the sudden drift detection method, and very often sudden drifts are mistakenly detected as gradual drifts. Thus, users need to check for sudden and gradual drifts separately, and then decide which drifts are relevant in a particular setting.


This article proposes an integrated and automated business process drift detection method addressing the above limitations. 
The core idea of the proposed method is to perform statistical tests over the distributions of \emph{partially ordered runs} observed in two consecutive time windows, where a run represents a set of traces that are equivalent to each other modulo a concurrency relation between the event types (activities) in the process. By re-sizing the window adaptively, the method strikes a trade-off between accuracy and drift detection delay while scaling up.

The basic method presented in this article detects sudden drifts. 
This method is then extended to detect gradual drifts based on the assumption that a gradual drift is delimited by two consecutive sudden drifts. To detect the location of gradual drifts, a post-processing step is applied wherein a separate statistical test is applied to the sets of runs observed between consecutive sudden drifts in order to determine if these sudden drifts represent separate changes, or if they define the start and end of a single gradual drift. Thus, the detection of sudden and gradual drifts is integrated, so that given an event log, 
the method returns a set of sudden drift points and a set of gradual drift intervals.

The proposed drift detection methods have been implemented and evaluated in terms of accuracy, detection delay and scalability, and compared against existing methods using a battery of synthetic and real-life event logs.


 

This article is an extended and revised version of a conference paper~\cite{Maaradji2015}. With respect to the conference version, the extensions include the definition, implementation and evaluation of the gradual drift detection method, including its comparison with an existing baseline; an extended evaluation for the basic method for sudden drift detection; a discussion of threats to validity; as well as a more extensive comparison with related work.

The article is structured as follows. Section~\ref{sec:background} discusses related work. Sections~\ref{sec:method} and \ref{sec:methodgrad} introduce respectively the sudden and gradual drift detection methods. Section~\ref{sec:implementation} presents the implementation while Sections~\ref{sec:evaluation1}-\ref{sec:evaluationReal} present the experimental evaluations. Finally, Section~\ref{sec:validity} discusses threats to validity, while Section~\ref{sec:conclusion} summarizes the contributions and discusses future work. 

\section{Background and Related Work}\label{sec:background}



The problem of business process drift detection is a variant of the problem of \emph{concept drift} detection in data mining~\cite{gama2014survey}. A concept drift refers to a change in the distribution of a variable. This change may occur suddenly or gradually~\cite{Minku2010}. A drift is \emph{sudden} if there is a time point such that the data observations before and after this point are generated by two different distributions, with no time overlap between them. Otherwise, the drift is said to be \emph{gradual}.

A large body of research work on concept drift in data mining has investigated the problem of identifying the time point when a drift occurs~\cite{gama2014survey}. 
For example, Klinkenberg et al.~\cite{klinkenberg2000detecting} detects drift by minimizing the error of a Support Vector Machine learner. 
Baena-Garcia et al.~\cite{baena2006early} proposed a method for fast drift detection based on monitoring the distance between classification errors. 
Ienco et al.~\cite{ienco2014change} introduced a technique for categorical data stream which requires at least two predictor variables though. 
In order to discard old data, different forgetting mechanisms has been proposed such as adaptive window technique~\cite{kifer2004detecting,adwin}, or fading factor~\cite{gama2009issues}.
Generally speaking, we observe that methods developed in this context deal with simple data structures (e.g.\ numerical or categorical variables and vectors thereof), while in business process drift detection we seek to detect changes in more complex structures, specifically behavioral relations between tasks (concurrency, conflict, loops). Thus, methods from the field of concept drift detection in data mining cannot be readily transposed to business process drift detection. 

Existing methods for business process drift detection are based on the idea of extracting \emph{patterns} (a.k.a.\ features) from traces. 
Bose et al.~\cite{bose2011handling,bose2014dealing} propose a method to detect process drifts based on statistical testing over feature vectors. This method is however not automated. Instead, the user is asked to identify the features to be used for drift detection, implying that the user has some apriori knowledge of the characteristics of the drift.  
If the user is unaware of the type of drift present in the log, they have to select all combinations of activities as the feature space, which is extremely computing-intensive.
Furthermore, given the types of features supported, this method is unable to identify certain types of drifts such as inserting a conditional branch or a conditional move. Finally, this method requires the user to set a window size for drift detection. Depending on how this parameter is set, some drifts may be missed.
This latter limitation is partially addressed in a subsequent extension~\cite{Martjushev2015}, which introduces a notion of \emph{adaptive window}. The idea is to increase the window size until it reaches a maximum size or until a drift is detected. This latter method requires that the user sets a minimum and a maximum window size. If the minimum window size is too small, minor variations (e.g.\ noise) may be misinterpreted as drifts. Conversely, if the maximum window size is too large, the execution time is affected and some drifts may go undetected. 

In ~\cite{Martjushev2015}, Martjushev et al. additionally propose a method for detecting gradual drifts. The idea is to apply their sudden drift detection technique by introducing a gap between the two sliding windows. This gap represents the gradual drift interval, whose maximal size needs to be manually initialized by the user. 
This design leads to a confusion between sudden and gradual drifts.
For example, when testing this method with logs containing only sudden drifts, the method detected theses drifts as gradual ones.




Accorsi et al. \cite{accorsi2012discovering} propose a drift detection method based on trace clustering. 
Similar to Bose et al.~\cite{bose2011handling,bose2014dealing}, this method heavily depends on the choice of window size, such that a low window size leads to false positives while a high window size leads to false negatives (undetected drifts), as drifts happening inside the window go undetected. 
In addition the method is not designed to deal with loops, and may fail to detect types of changes that do not significantly affect the distances between activity pairs in a trace, e.g.\ changes involving an activity being always executed before the drift, and sometimes skipped after the drift.

Carmona et al. \cite{carmona2012online} propose another drift detection method based on an abstract representation of the process as a polyhedron. This representation is computed for prefixes extracted from a random sample of the earlier traces in the log. The method checks the fitness of subsequent prefixes of traces against the constructed polyhedron. If a large number of these prefixes do not lie in the polyhedron, a drift is declared. To find a second drift after the first one, the entire detection process has to be restarted, thus hindering on the scalability of the method. In experiments we conducted with the logs used in Sections~\ref{sec:evaluation1} and~\ref{sec:evaluationGrad}, the implementation of this method took hours to complete. Another drawback of this method is its inability to pinpoint the exact moment of drift.

Burattin et al.~\cite{burattin2014control} adapt an automated process discovery method, namely the Heuristics Miner, to handle incremental updates to the discovered process model as new events are observed. Our proposal is complementary to the latter as it allows drifts to be detected accurately and efficiently, and can be used as an oracle to identify points in time when the process model should be updated. 



In a subsequent work, Burattin et al. \cite{Burattin2015} present a method for online discovery  of process models captured as a set of constraints between events. These constraints are formulated in the DECLARE language based on Linear Temporal Logic (LTL). The work explores different streaming techniques to update the LTL constraints incrementally. In this setting, any change in the extracted constraints over time may be considered as a drift. However, no statistical support is provided to analyze the significance of changes in order to discern transient changes from drifts. Moreover, DECLARE lacks support for explicitly capturing the concurrency relation, leading to some changes being potentially missed. Finally, the exact position of the changes is not reported.  

In a separate study~\cite{ostovar2016detecting}, we addressed the problem of sudden drift detection in business processes with high variability, i.e.\ processes with a high proportion of distinct traces relative to the total number of traces. Drift detection in such processes requires specialized methods because the set of possible distinct traces is so large that previously unobserved traces keep appearing continuously, though this does not necessarily imply that the process behavior has drifted. In~\cite{ostovar2016detecting}, we tackle this issue by performing statistical tests not over the distributions of runs, but rather over the distributions of behavioral relations between pairs of events. This method is designed to work with streams of events in online settings. However, it only deals with sudden drifts, not with gradual ones. 

In a subsequent study \cite{ostovar2017characterizing}, we extended the method in \cite{ostovar2016detecting} to characterize the identified drifts. Specifically, we perform a statistical test to measure the statistical association between the drift and the distributions of behavioral relations extracted from the event stream before and after the drift. We then rank these relations based on their relative frequency change, and match them with a set of predefined process change templates. The best-matching templates are then reported to the user via natural language statements (e.g.\ ``Before the drift, activity E preceded F, while after the drift they are in parallel''). 

This article is an extended and revised version of a conference paper~\cite{Maaradji2015}. While the conference paper focuses on sudden drift detection, this article extends the scope of the study to cover both sudden and gradual drifts in an integrated manner.



\section{Sudden drift detection method}
\label{sec:method}

%



The problem of business process drift detection can be formulated as follows: \emph{identify a time point when there is a statistically significant difference between the observed behavior before and after this point}. To turn this statement into a decision procedure, we first need to define what we mean by a \emph{difference in the observed behavior}. In other words, we need to define \emph{when are two processes the same?}~\cite{Hidders2005}. A number of equivalence notions have been proposed to address this question. One widely accepted notion of process equivalence is \emph{trace equivalence}: two processes are the same if they have the same set of traces, thus they are different if their set of traces exhibits a (statistically significant) difference. This trace-based representation does not capture concurrency. As a result, any significant variation in the frequency of relative ordering of two activities that are anyways in parallel will be treated as a drift. For example, if two activities \textsf{b} and \textsf{c} are in parallel, any significant variation in the frequency of occurrence of \textsf{b followed by c} vs.\  \textsf{c followed by b} gives rise to a drift, even though the parallel relation between these activities still holds. 
From this perspective, a more suitable approach is to reason in terms of \emph{partially ordered runs} (a.k.a.\ configurations) of a process, where concurrency is explicitly captured.  For example, the two traces \textsf{abcd} and \textsf{acbd} characterize the process where \textsf{a} is followed by the block \textsf{b} and \textsf{c} in parallel, and these are followed by \textsf{d}. In a run-based representation, only one run is needed to represent both traces: the run where \textsf{a} is followed by \textsf{b} and \textsf{c} in parallel and these are followed by \textsf{d}. As business processes typically contain concurrent activities, we opt for a run-based representation of logs and thus a notion of run-equivalence, known as \emph{configuration equivalence} or \emph{pomset equivalence}~\cite{vanGlabbeek1990}.

\begin{figure}[hbt]
\centering
\captionsetup{justification=centering,margin=1cm}
\includegraphics[width=0.95\linewidth]{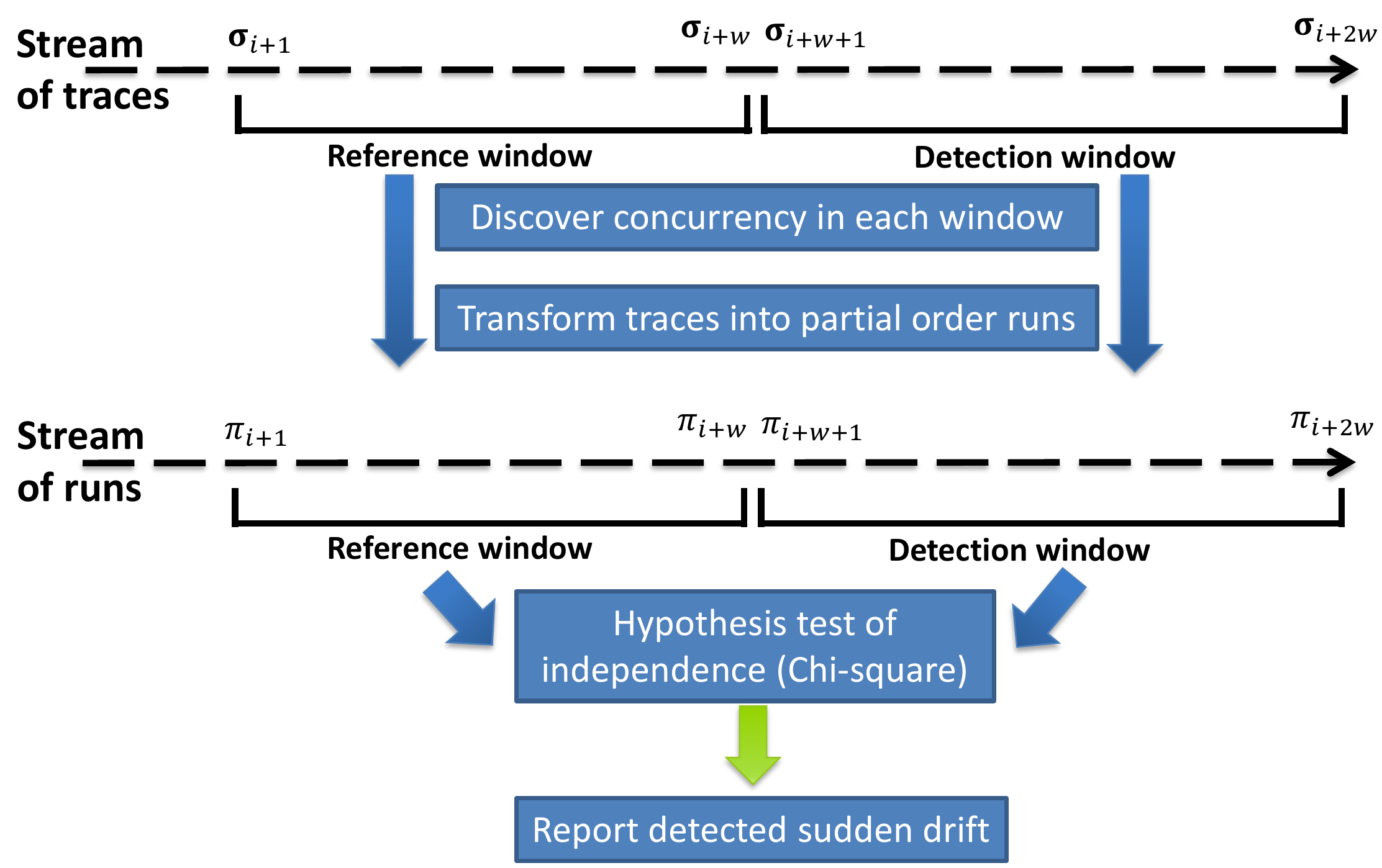} 
\caption{Overview of sudden drift detection method}
\label{fig:overviewsudden}
\vspace{-\baselineskip}
\end{figure}

Given the above, we map the problem of process drift detection to that of finding a time point such that the set of runs before this point is statistically different from the set of runs after (for a given observation window size). This formulation leads to the two-staged approach outlined in Figure~\ref{fig:overviewsudden}. In the first stage, we divide the recently observed traces into two windows: the detection window (most recent traces) and the reference window (older traces). For each window, we discover a set of concurrency relations between the event labels (activities) observed in the window and then transform the set of traces that \emph{completed} in each time window into a set of runs. In the second stage, we apply statistical testing to find significant differences between the sets of runs of adjacent time windows. 
In a preprocessing step, we replay completed traces of a given event log in their temporal order to produce a stream of traces, that is used as input to our method.
The next two sub-sections discuss these two stages in turn, while the third sub-section discusses how the window size is adapted dynamically.


\subsection{From event logs to partial order runs}

An event log consists of a set of traces, each capturing the sequence of events observed for a given case of the process, ordered by timestamp. 
For example using an simplified notation,  $L=\left [ \sigma_1^2,\sigma_2^3 \right ]$, where $\sigma_1=\left<a,b,c,d\right>$ and $\sigma_2=\left<a,c,b,d\right>$, defines a log containing 5 traces (two occurrences of $\sigma_1$ and three of $\sigma_2$) and a total of 20 events. Formally: 

\begin{definition}[Event log, Trace]
Let $L$ be an \emph{event log} over the set of labels $\mathcal{L}$, i.e. $L \in \mathbb{B}(\mathcal{L}^*)$. Let $E$ be a set of event occurrences and $\lambda : E \to \mathcal{L}$ a labeling function.
An \emph{event trace} $\sigma \in L$ is defined in terms of an order $i \in [0,n-1]$ and a set of events $E_\sigma \subseteq E$ with $|E_\sigma| = n$ such that $\sigma = \left<\lambda(e_0), \lambda(e_1), \dots, \lambda(e_{n-1}) \right>$.
\label{def:log-trace}
\end{definition}

A trace defines a \emph{total order} between a  set of events. 
If we know that some of these events are related via concurrency relations, we can transform a trace into a partially ordered run by ``breaking'' the total order between concurrent events in the trace. For simplicity, in the following we use the notion of \emph{Alpha concurrency} defined in~\cite{AalstWM04} to identify concurrent pairs of events. 


\begin{definition}[$\alpha$ concurrency]
Let $L$ be an event log over the set of event labels $\mathcal{L}$ and $\sigma \in L$ be a log trace.
A pair of tasks with labels $A, B \in \mathcal{L}$ are said to be in
\emph{$alpha$-directly precedes relation}, denoted $A \prec_{\alpha(L)} B$, iff
there exists a trace \mbox{$\sigma = \langle \lambda(e_0), 
\lambda(e_1),\ldots,\lambda(e_{n-1}) \rangle$} in $L$, such that $A =
\lambda(e_i)$ and $B = \lambda(e_{i+1})$. We say that a pair of tasks $A, B \in
\mathcal{L}$ are \emph{$\alpha$-concurrent}, denoted $A \parallel_{\alpha(L)} B$, iff \mbox{$A \prec_{\alpha(L)} B \land B \prec_{\alpha(L)} A$}.
\label{def:ar}
\end{definition}

Note that $\alpha$-concurrency is a symmetric relation and that it applies to event labels and not over event occurrences. For instance, we can identify that $\text{b} \prec_\alpha \text{c}$ from trace $\sigma_1=\left<a,b,c,d\right>$, and $\text{c} \prec_\alpha \text{b}$ from trace $\sigma_2=\left<a,c,b,d\right>$. Therefore, $b$ and $c$ are considered to be parallel, noted $b \parallel_\alpha c$.

It is possible to use more sophisticated concurrency relations such as the $\alpha^+$~\cite{de2003workflow}, $\alpha^{++}$~\cite{wen2007mining}, or the one proposed in \cite{Cook1998}. 
The choice of concurrency relations does not fundamentally alter our approach.
To abstract away from any specific approach to extract concurrency relations, we assume in the sequel that we are given as input a function $\chi$ (herein called a \emph{concurrency oracle}), which given an event log $L$, returns a symmetric concurrency relation $\parallel_{\chi(L)}$ between event labels in the log. In particular, if we are given the $\alpha$ concurrency oracle as input, then $\parallel_{\chi(L)}\ = \parallel_{\alpha(L)}$.

Given the above, Definition~\ref{def:trace2run} describes how a
trace is transformed into a partially ordered run given a concurrency oracle. 

\begin{definition}[Transformation of a trace into a run]
\label{def:trace2run}
Let $L$ be an event log over the set of event labels $\mathcal{L}$ and
$\parallel_{\chi(L)}$ be the concurrency relation provided by an oracle $\chi$ applied to $L$.
Let $E$ be a set of event occurrences, $\lambda : E \to \mathcal{L}$ a labeling function.
We say that event $e_{i}$ \emph{directly precedes} event $e_{i+1}$, denoted
$e_i\ \lessdot\ e_{i+1}$, iff there exists a trace $\sigma = \langle \lambda(e_0), \ldots, $
$ \lambda(e_{n-1}) \rangle$ in $L$ with an order  $i \in [0,n-1]$.
Therefore, the tuple $\pi = \langle E_\pi, \leq_\pi, \lambda_\pi \rangle$ is the \emph{partially ordered run} corresponding to trace $\sigma$, induced by the concurrency relation $\parallel_{\chi(L)}$ and the directly precedes relation $\lessdot$, where:
\begin{compactitem}
\item $E_\pi$ is the set of events occurring in $\sigma$,
\item $\leq_\pi$ is the causality relation defined as ${\leq_\pi\ = E_\pi^2 \cap (\lessdot^+ \setminus \parallel_{\chi(L)})^*}$,
and
\item $\lambda_\pi : E_\pi \to \mathcal{L}$ is a labeling function, i.e. $\lambda_\pi = \lambda|_{E_\pi}$.
\end{compactitem}

\noindent
We write $\Pi_\chi(L)$ to denote the set of all partially ordered runs induced by $\parallel_{\chi(L)}$ over the set of traces in $L$.
\label{def:por}
\end{definition}

\begin{wrapfigure}{r}{0.35\linewidth}
\vspace*{-2\baselineskip}
\centering
\begin{tikzpicture}[->,thick,font=\large]
\matrix[row sep=2.5mm, column sep=0.2mm, ampersand replacement=\&]{
\& \node(a) {a}; \\
\\
\node(b)   {b};\&\&\node(c) {c};\\
\\
\&\node(d) {d}; \\
};
\path (a) edge (b) (c) edge (d);
\path (b) edge (d);
\path (a) edge (c);
\end{tikzpicture}
\caption{Sample run ($\pi_1$)}
\label{fig:run1}
\vspace{-0.9\baselineskip}
\end{wrapfigure}
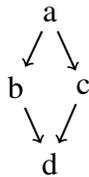

Let us consider event log $L$ and apply the definition step-by-step. We first compute the directly-precedes relationship $\lessdot$ by representing the sequencing captured by the event traces, resulting in set $\{ (a_{\sigma_1} , b_{\sigma_1}),(b_{\sigma_1} , c_{\sigma_1}),(c_{\sigma_1} , d_{\sigma_1}),(a_{\sigma_2} , c_{\sigma_2}),(c_{\sigma_2} , b_{\sigma_2}), (b_{\sigma_2} , d_{\sigma_2}) \}$. Second, we compute the (irreflexive) transitive closure $\lessdot^+$ by adding to the previous set the following new relations: $\{ (a_{\sigma_1} , c_{\sigma_1}),(b_{\sigma_1} , d_{\sigma_1}),(a_{\sigma_1} , d_{\sigma_1}),(a_{\sigma_2} , b_{\sigma_2}),(c_{\sigma_2} , d_{\sigma_2}),\\(a_{\sigma_2} , d_{\sigma_2}) \}$. Third, we compute the concurrency relation $\parallel_{\chi(L)}$, and obtain $\{ (b,c) \}$. Fourth, we compute the causality relation $\leq_{\pi_1}$ for the run $\pi_{1}$ corresponding to trace $\sigma_{1}$ by computing set $\lessdot^+ \setminus \parallel_{\chi(L)}$, which leads to removing relation $(b_{\sigma_1} , c_{\sigma_1})$. Similarly, we remove  $(c_{\sigma_2} , b_{\sigma_2})$ for $\leq_{\pi_2}$ for run $\pi_{2}$ from $\sigma_{2}$. Finally, we remove the unnecessary transitive relations $(a_{\sigma_1} , d_{\sigma_1})$ for $\pi_{1}$ and $(a_{\sigma_2} , d_{\sigma_2})$ for $\pi_{2}$.

Using only event labels as a simplified representation, the result of this transformation applied on $\sigma_1= \left<a,c,b,d\right>$ is the run $\pi_1$ defined by the following causality relation $\leq_{\pi}=\{(a , b), (a , c) , (b , d), (c , d) \}$ implicitly inferring that $b \parallel_{\chi(L)} c$ (cf. Figure \ref{fig:run1}).  
In the simplified notation, each trace in $L$ is transformed to the exact same run represented by $\pi$.
In the remainder of the paper, a \emph{run} always refers to its simplified representation.

Armed with these concepts, we can transform a set of traces into a set of runs by first calculating the alpha relationships, and then transforming each trace into a run as outlined above. Below, we show how sets of runs  observed in two consecutive time windows are analyzed in order to detect sudden drifts.




\vspace{-.5\baselineskip}
\subsection{Statistical test and oscillation filter}\label{sec:chi_square}\label{sec:oscillation}

\begin{figure}
\centering
\captionsetup{justification=centering,margin=1cm}
\includegraphics[width=\linewidth]{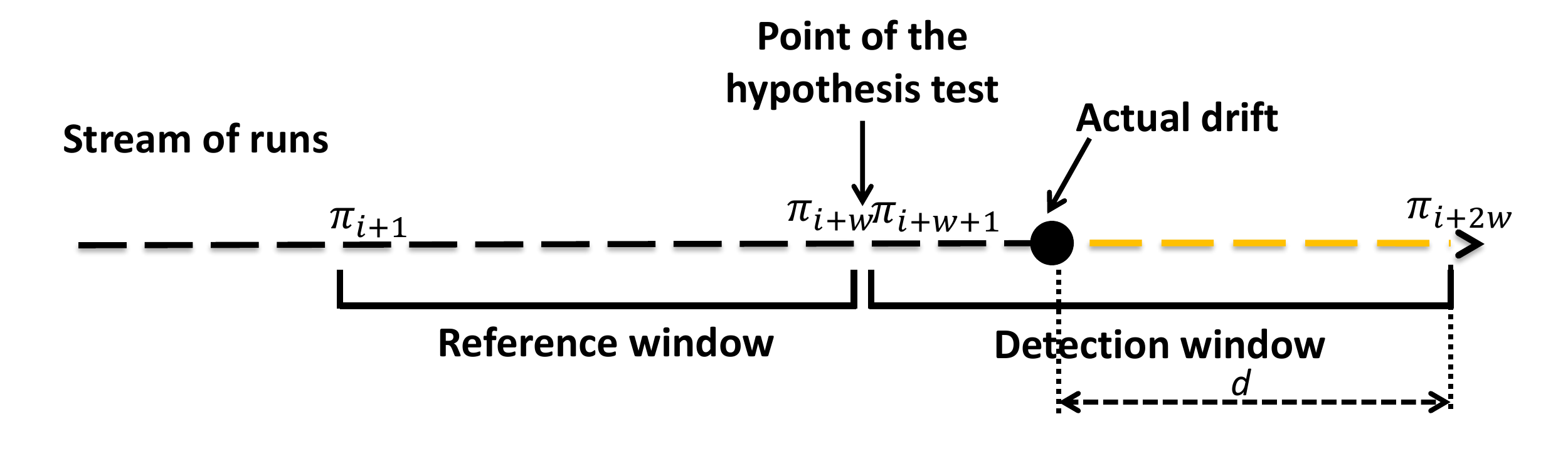}
\vspace{-6mm} 
\caption{Statistical test over two sliding windows}
\vspace{-\baselineskip}
\label{fig:2win}
\end{figure}

In order to detect a drift in a stream of runs, 
we monitor any statistically significant change in the distribution of the most recent runs. 
This test is done on two populations of the same size $w$ built from the most recent runs in the stream. 
Basically the most recent runs are divided into a \emph{reference} (less recent) and a \emph{detection} (more recent) populations, forming together the \emph{composite window}. Then, we evaluate the statistical hypothesis of whether the reference and detection populations of runs are similar. 
Figure \ref{fig:2win} depicts a reference and a detection window over a stream of runs. 
For every new run in the stream, we slide both windows to the right in order to read the new run and perform a new statistical test. We keep iterating this process as long as there are new runs observed in the stream.
 
Since there is no a-priori knowledge of the run distributions (and their parameters) within the reference and detection windows population, we apply a non-parametric hypothesis statistical test. Moreover, given that an observation of the statistical variable is a run, the statistical test has to be applicable to a categorical variable. For these reasons, we selected the Chi-square test of independence between two variables. 
The goal of a two-variable Chi-square test is to determine whether the reference variable and detection variable are similar. The reference variable (resp., the detection variable) is represented by the observations from the reference window (resp., the detection window). A contingency matrix is built to report the frequencies of each distinct run in each window. The Chi-square test is performed on this contingency matrix which return the significance probability (the \emph{P-value}). 
A drift is detected when the P-value is less than the significance level (the threshold), and localized at the point of juxtaposition of the reference and detection windows. The value of the threshold is set to the typical value of the Chi-square statistical test, which is 0.05~\cite{nuzzo2014statistical}. 

The \emph{delay} $d$ shown in Figure \ref{fig:2win} is a notion from concept drift in data mining \cite{ICML05martingale}. It is not the distance between the actual drift and the location where the drift is detected. Rather, it indicates how long it takes for the statistical test to detect the drift after it has occurred, and is measured as the number of runs between the drift and the end of the detection window. 

Since any statistical test is subject to sporadic stochastic oscillations, we introduced an additional filter, namely the \emph{oscillation filter}, to discard abrupt stochastic oscillations in the P-value. An abrupt stochastic oscillation is caused by the noise present in the event log, e.g.\ in the form of infrequent events or data gaps. Accordingly, we detect a drift only if a given number $\phi$ of successive statistical tests have a P-value less the typical threshold for the statistical test. In other words, a persistent P-value under the threshold is much more reliable than a sparse value happening abruptly. 
Tests that we carried out using the \textit{simple moving average (SMA)} smoothing technique on the P-value curve, led to similar results. More sophisticated approaches to filter out stochastic oscillations are however available, e.g.\ from the financial domain \cite{murphy1999technical}, and could be used instead.


The only independent parameter in the proposed method is the window size $w$. Below we discuss how to automatically adjust this parameter as new runs are observed in the stream. 

\vspace{-.5\baselineskip}
\subsection{Adaptive window}
\label{sec:adwin}

As discussed in Section \ref{sec:background}, the choice of window size is critical in any drift detection method as a small window size may lead to false positives while a large one may lead to false negatives as well difficulty in locating the exact point of the drift. Our method strikes this trade-off by adapting the window size in order to have a more reliable statistical test. It is inspired by \cite{adwin}, where the authors provide rigorous guarantees on the performance of the adaptive window technique. 

The underpinning observation of our method is that if the variation within the composite window is high (resp. low), we need more (resp. less) observations to assert that the distribution of runs in the detection window is statistically different from the distribution of runs in the reference window. Accordingly, we increase the composite window size if we observe high variability, and decrease it if we observe low variability.

The variability (a.k.a.\ variation) of a statistical variable is a measure of the dispersion of a set of random observations of the variable. 
In the case of a categorical variable with a large number of possible values (e.g. the set of possible runs of a process), higher dispersion entails a higher number of observed distinct values relative to all observations.
Accordingly, we measure variability as the ratio between the number of distinct runs and the total number of the runs in a composite window. 

When we slide the composite window, the total number runs in the new window and in the old window is initially the same. What can vary between these two consecutive sliding windows is the number of distinct runs. In order to keep the variability constant between these two consecutive sliding windows, we adjust the size of the new window proportionally to the change in the number of distinct runs -- if the number of distinct runs goes down (up), the size of the window will decrease (increase).
Formally, given two consecutive composite windows $T_{1}$ and $T_{2}$, the $evolution~ratio$ between $T_{2}$ and $T_{1}$ is defined as the ratio between the number of distinct runs in $T_{2}$ and the number of distinct runs in $T_{1}$. An  $evolution~ratio$ of 1 means that there is no evolution in the variability between $T_{1}$ and $T_{2}$. An $evolution~ratio$ less than 1 means that there is less variation in $T_{2}$ relative to $T_1$, whereas an $evolution~ratio$ greater than 1 means the opposite. 

The composite window size is adjusted based on the $evolution~ratio$  every time that it is shifted forward to incorporate a new run in the stream, specifically: $nextWindowSize = currentWindowSize \cdot evolutionRatio$. To initialize the procedure, we start with a user-defined window size. 
If the user does not provide an initial window size, the method uses the minimum size required by the statistical test. For instance, for the chi-square test, no more than 5\% of the frequencies in the contingency table are allowed to be less than five.

The complete sudden drift detection algorithm and its time complexity analysis can be found in the online Appendix. 


\section{Gradual drift detection method}
\label{sec:methodgrad}

A gradual drift is intuitively defined as \emph{a change that happens progressively from one underlying process behavior to another one}. In contrast to a sudden drift where an abrupt change occurs at a time point, a gradual drift is characterized by a transition time interval where both the prior and posterior process behaviors overlap. Throughout this transition \emph{interval}, the prior process behavior fades-out whereas the posterior one fades-in. 


The proposed gradual drift detection method relies on the assumption that a gradual drift is delimited by two consecutive sudden drifts discernible by the method described in Section \ref{sec:method}, such that the distribution of runs in the interval between these two drifts is a mixture of the distributions of runs before the first drift and after the second drift.
The reasoning behind this assumption is that there is a point in time where the new process behavior starts to fade-in (leading to the first sudden drift), and a point in time where the old process behavior fades-out completely (leading to the second drift).
By virtue of the definition of drift (i.e.\ a change in the behavior of the process), the distribution of runs before the start of the gradual drift (i.e.\ before the first sudden drift) is different from the distribution of runs after the end of the gradual drift (i.e.\ after the second sudden drift). 
In the interval between the two sudden drifts (herein the \emph{transition interval}), the old and the new behavior co-exist. This leads us to assume that the distribution of runs in the transition interval follows a statistical distribution that is a mixture of the distribution before the first sudden drift and of the distribution after the second sudden drift.

Accordingly, gradual drifts are detected by post-processing the output of the sudden drift detection algorithm. Specifically, each two consecutive sudden drifts are statistically analyzed in order to determine whether the behavior in the interval between them is a mixture of the behaviors before and after them.

\vspace{-.5\baselineskip}
\subsection{Statistical testing for gradual drifts}

We seek to test if the distribution of runs in-between two consecutive sudden drifts (the transition interval) is a mixture of the distribution of runs before the first sudden drift and the distribution of runs after the second sudden drift. 
We further assume that this ``mixture'' is linear, meaning that the distribution in the interval can be written as a linear combination of the distributions before and after the interval.
Hence, we seek to find weights $ \left( x_0, y_0\right) $, such that the weighted addition of the distributions before and after the transition interval fits the distribution observed during the interval.
The fitness between the linear mixture and the actually observed distribution 
is asserted if we are able to confirm the null hypothesis of the Chi-square goodness-of-fit statistical test. This test is designed to determine if a set of observations is consistent with a hypothesized distribution, in this case the linear mixture.

If no tuple $ \left( x_0, y_0\right) $ can be found for which the null hypothesis is confirmed, we conclude that the distribution in the transition interval is not consistent with any linear mixture of the distributions of the surrounding fragments (no gradual drift). On the other hand, a gradual drift is declared if we can find at least one tuple $ \left( x_0, y_0\right) $ for which the null hypothesis is confirmed. 

Since we do not have the full statistical distribution of runs over a given time interval, we use the histogram of runs as a proxy. Histograms have been shown to be a fair approximation of a categorical variable's statistical distribution~\cite{Guha2006} and they have already been used for drift detection in data mining \cite{sebastiao2017fading}.

For every two consecutive sudden drifts, the gradual drift detection method maintains the histograms of populations of runs corresponding to the following intervals: (\emph{i}) directly before the start of the first sudden drift, (\emph{ii}) between the two drifts, and (\emph{iii}) directly after the second drift.
We then apply a Chi-square goodness-of-fit statistical test on these histograms to assert if the process behavior within the interval is a mixture of the two process behaviors before and after it, as formalized below.

\begin{definition}[Gradual drift detection]

Given:
\begin{compactitem}
\item two consecutive sudden drifts $ D_{k-1} $ and  $ D_{k} $, delimiting the populations of runs  $P_{k-1},  P_{k}, P_{k+1}$, and
\item the histograms $ H_{k-1}$, $H_{k}$ , and $H_{k+1}$  representing the population $P_{k-1}$,  $P_{k}$, and $P_{k+1}$ respectively
\end{compactitem}
A gradual drift delimited by $ D_{k-1} $ and  $ D_{k} $ has occurred if:
\begin{compactitem}
\item $ \exists \left( x_0, y_0\right) \in \mathcal{R}$, and $\hat{H_{k}}$ a histogram where $\hat{H_{k}} = x_0H_{k-1}  + y_0H_{k+1}$, and
\item the null hypothesis $ \hat{H_{k}} \approx H_{k} $ can not be rejected by the Chi-square goodness-of-fit test.
\end{compactitem}
 

\label{def:graddrift}
\end{definition}

\begin{figure}[hbt!]
\centering
\captionsetup{justification=centering,margin=1cm}
\includegraphics[width=\linewidth]{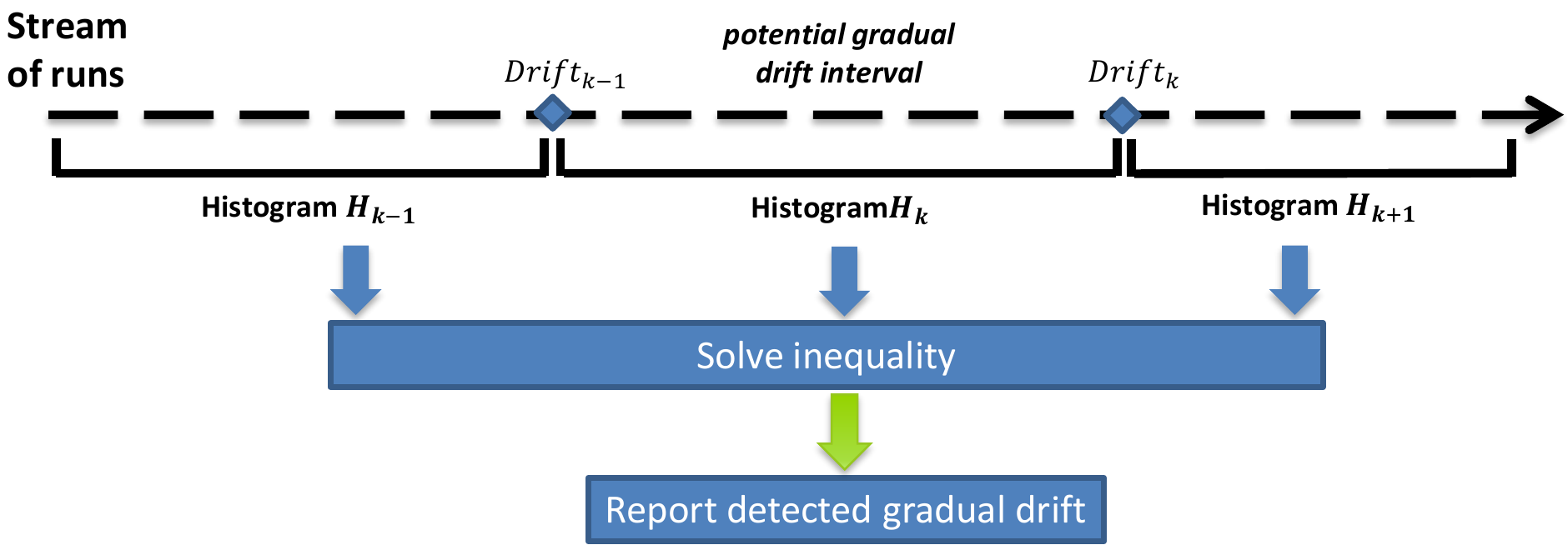}
\caption{Overview of gradual drift detection method}
\label{fig:graddrift}
\end{figure}

The population $ P_{k} $ of runs within the candidate gradual drift interval (i.e.\ the interval between two sudden drifts) is represented by histogram $ H_{k} $, whereas histograms $ H_{k-1}  $ and $ H_{k+1} $ represent run distributions before and after the candidate gradual drift as depicted in Figure~\ref{fig:graddrift}. If a mixture distribution of $ H_{k-1}  $ and $ H_{k+1} $ is possible, represented here by $\hat{H_{k}}$, and if this mixture $\hat{H_{k}}$ is similar to $H_{k} $, we can state that $\hat{H_{k}}$ is likely to have generated the observation $ P_{k} $ and thus $ P_{k} $ is likely to be a population of a gradual drift interval.

According to Definition~\ref{def:graddrift}, we are looking for a tuple $\left( x_0, y_0\right)$ such that we can statistically validate the hypothesis that $\hat{H_{k}} $ is a mixture of $ H_{k-1} $ and $ H_{k+1} $ with weights $x_0$ and $y_0$ respectively.
Based on the \textit{Chi-square goodness-of-fit test}, this hypothesis is not rejected with confidence level $\left( 1- {Tr}_{Chi}\right)$ if: 
\vspace{-.3\baselineskip}
\begin{equation}\sum_{i=1}^{df+1} \frac{({H_{k}}_i - {\hat{H_{k}}}_i)^2} {{\hat{H_{k}}}_i} < {\chi^2}_{ df} \left( {Tr}_{Chi} \right) 
\label{eq:1}
\end{equation}
\vspace{-.5\baselineskip}

where $ {\chi^2}_{ df} \left( {Tr}_{Chi} \right) $ is the Chi-square critical value given as the point where the complement of the chi-square cumulative distribution function equals ${Tr}_{Chi}$. 
The Chi-square critical value is retrieved from the Chi-square table based on the significance level (here ${Tr}_{Chi}$) and the degrees of freedom of the test (here~$df$). 
The degrees of freedom are equal to the number of categories (distinct runs) minus one. Replacing $ \hat{H_{k}} $ by $  xH_{k-1}  + yH_{k+1} $ in Inequality~\ref{eq:1} results in:
\vspace{-.3\baselineskip}
\begin{equation}\label{inequality} 
\sum_{i=1}^{df+1} \frac{\left({H_{k}}_i - \left( x{H_{k-1}}_i  + y{H_{k+1}}_i  \right) \right)^2} {x{H_{k-1}}_i  + y{H_{k+1}}_i} < {\chi^2}_{ df} \left( {Tr}_{Chi} \right) 
\end{equation}
\vspace{-.5\baselineskip}

We use a non-linear programming (NLP) solver to try to solve Inequality~\ref{inequality}. If the NLP solver returns a solution $\left( x_0, y_0\right)$ satisfying Inequality~\ref{inequality}, then the null hypothesis is valid. The validity of the null hypothesis confirms that $ H_{k} $ conforms to the distribution $ \hat{H_{k}} $. This means that with a confidence level $\left( 1- {Tr}_{Chi}\right)$, $ P_{k} $ has been generated by $ \hat{H_{k}} $, itself a mixture of $ H_{k-1}  $ and $ H_{k+1} $. Hence, a gradual drift is declared. Note that the returned solution $\left( x_0, y_0\right)$ indicates that the relative weight of the histogram $ H_{k-1} $ (resp. $ H_{k+1} $) in the gradual drift is $ x_0/(x_0+ y_0)$ ( resp. $ y_0/(x_0+ y_0)$).

%

Figure~\ref{fig:graddrift} provides an overview of the gradual drift detection approach. The detailed algorithm 
and its time complexity analysis are given in the online Appendix. 

\vspace{-.5\baselineskip}
\subsection{Example}

To illustrate the gradual drift detection method, let us assume we have an event log of 2000 traces with a \emph{linear gradual drift} of 400 traces. In the first step, Algorithm~\ref{alg:sdrift} detects two sudden drifts at the start and end of the gradual drift, say drift one at trace index 820, and the second drift at trace index 1250. 
Second, we extract the three histograms delimited with these two consecutive drifts, namely $ H_1, H_2, H_3$ corresponding to the trace ranges [0-819], [820-1249], and [1250-1999] respectively.
Third, let us assume the number of distinct runs in these three histograms is 41. Consequently, the degrees of freedom of the Chi-square statistical test is 40. The fourth step is to fetch the Chi-square critical value, corresponding to the degrees of freedom $df = 40$ and a significance level ${Tr}_{Chi} = 0.05$, which corresponds to 55.758 (retrieved from the Chi-square test table).
Finally, based on the Chi-square critical value and the histograms, we populate the inequality \ref{inequality}, which is then solved with the NLP solver. Let us assume the solver returns the solution $(x_0,y_0) = (0.5, 0.5)$. Finding a solution for the inequality actually means the null hypothesis can not be rejected by the Chi-square test. Hence, the detected two consecutive sudden drifts delimit a gradual drift. The solution $(x_0,y_0) = (0.5, 0.5)$ means that the histogram $ H_2 $ is an equally balanced combination of $H_1$ and $H_3$. An \emph{a posteriori} Chi-square test of the null hypothesis that $ H_2 $ conforms to $ 0.5H_1 + 0.5 H_3$ results in a significance probability (P-value) above $0.05$.


%

%
%
%

\section{Implementation}\label{sec:implementation}
We implemented the two methods for drift detection in an open-source software tool called \emph{ProDrift}, available as a plugin of the Apromore online platform,\footnote{Available at \url{http://apromore.org}} and a standalone Java command-line tool.\footnote{Available at \url{http://apromore.org/platform/tools}} 
ProDrift relies on JOptimizer\footnote{Available at \url{http://www.joptimizer.com}} as the NLP solver. We used JOptimizer because it is an efficient, open-source NLP solver developed in Java.

%

When using the ProDrift plugin in Apromore, users can import an event log in XES or MXML format. 
Each new trace is used to dynamically update the alpha-relationships for each pair of events, and then transformed to a partial order run, resulting in a stream of runs. This stream of runs is then used as input for the statistical test. The detected sudden and gradual drifts are reported on a plot of the P-value curve, as well as in a list which includes the number of traces and timestamps in the log where the drifts were found, as well as the mean delay expressed as the number of traces read from the log. The user can then extract sublogs defined by each two consecutive drifts, representing ``stable'' process behaviors. These sublogs can be used for further analysis, e.g.\ for analyzing the evolution of a business process over time~\cite{Conforti2015} by relying on other plugins available in Apromore. 

\section{Evaluation of sudden drift detection on synthetic logs}\label{sec:evaluation1}

In this section we report on the evaluation of our basic method for detecting sudden drifts. First, we evaluate the impact of the various parameters (oscillation filter size, window size, and use of adaptive window) on the accuracy of the results. We do so by using a synthetic dataset, in order to test the method against different types of changes and inter-drift distances.
To assess accuracy we use two established measures in concept-drift detection in data mining \cite{ICML05martingale}, namely the \emph{F-score}, measured as the harmonic mean of recall and precision, and the \emph{mean delay}. The latter, computed as the average number of log traces after which a drift is detected, not only measures how late we detect the drift with regard to where it actually happens, but it also indicates how far in the log traces are read to be able to detect a drift.

Once we identify the best values for the parameters, we assess the accuracy of our method for various types of sudden drifts, and compare the results with the baseline method in \cite{bose2014dealing}. Finally, we report on time performance.

\vspace{-.5\baselineskip}
\subsection{Dataset generation}\label{sec:patterns}

%

\begin{figure*}[!htb]
\centering
\includegraphics[width=0.85\linewidth]{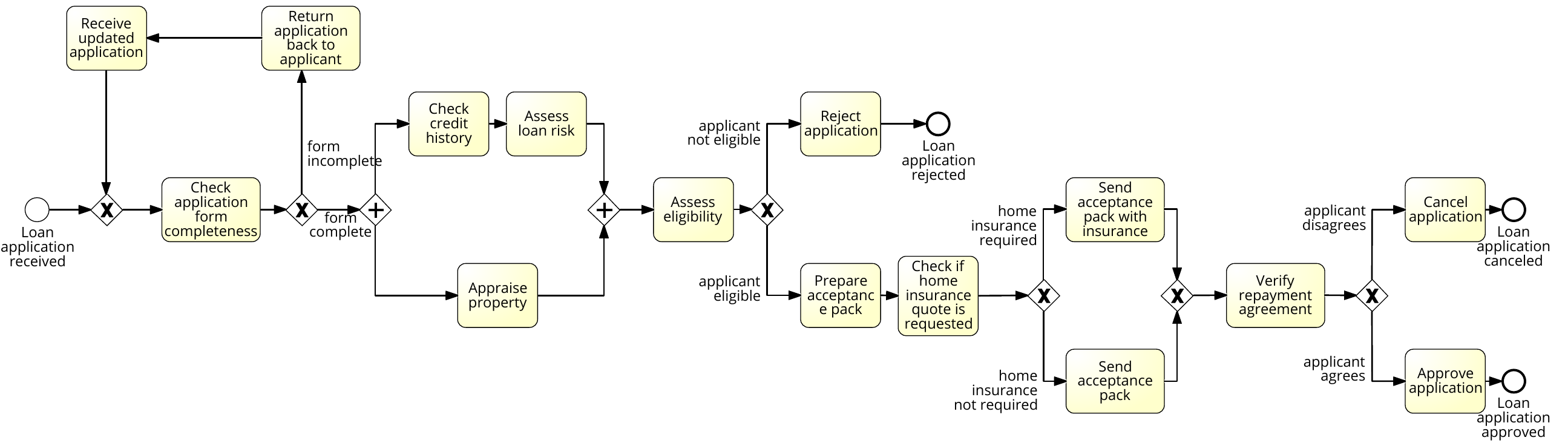} 
\caption{Base BPMN model of the loan application process}
\label{fig:Loan_baseline}
\vspace{-\baselineskip}
\end{figure*}

We generated a benchmark of 72 event logs by taking a textbook example of a business process for assessing loan applications \cite{Dumas2013} as a ``base'' model and altering it as outlined below. This model, illustrated in Figure \ref{fig:Loan_baseline}, has 15 activities, one start event and three end events, and exhibits different control-flow structures including loops, parallel and alternative branches. 

In order to assess the ability of our method to detect sudden drifts determined by different types of control-flow changes, we systematically altered the base model by applying in turn one out of twelve \emph{simple change patterns} described in \cite{Weber2008}.\footnote{Subprocess extraction and inlining patterns were excluded as they do not affect the process itself.} These patterns, summarized in Table~\ref{tab:patterns}, capture different change operations commonly identified in business process models, such as adding, removing or looping a model fragment, swapping two fragments, or parallelizing two sequential fragments. 

\begin{table}[h]
\centering\scriptsize{
\begin{tabular}{ | c | l | c | }
\hline
	\textbf{Code} & \textbf{Simple change pattern} & \textbf{Category} \\ \hline \hline
	re & Add/remove fragment & I \\ \hline
	cf & Make two fragments conditional/sequential & R \\ \hline
	lp & Make fragment loopable/non-loopable & O \\ \hline
	pl & Make two fragments parallel/sequential & R \\ \hline
	cb & Make fragment skippable/non-skippable & O \\ \hline
	cm & Move fragment into/out of conditional branch & I \\ \hline
	cd & Synchronize two fragments & R \\ \hline
	cp & Duplicate fragment  & I \\ \hline
	pm & Move fragment into/out of parallel branch & I \\ \hline
	rp & Substitute fragment  & I \\ \hline
	sw & Swap two fragments  & I \\ \hline
	fr & Change branching frequency  & O \\ \hline
\end{tabular}}
\caption{Simple control-flow change patterns}\label{tab:patterns}
\end{table}

To emulate more complex drifts, we organized the simple changes into three categories: Insertion (``I''), Resequentialization (``R'') and Optionalization (``O'') as shown in Table~\ref{tab:patterns}, so as to give rise to six possible \emph{composite change patterns} by randomly applying one pattern from each category in a nested way (``IOR'', ``IRO'', ``OIR'', ``ORI'', ``RIO'', ``ROI''). For example, the composite pattern ``IOR'' was obtained by first adding a new activity (``I''), then making this activity in parallel with an existing activity (``R'') and finally by putting the parallel block in a loop structure (``O'').

\begin{figure}[!htb]
\centering
\includegraphics[width=\linewidth]{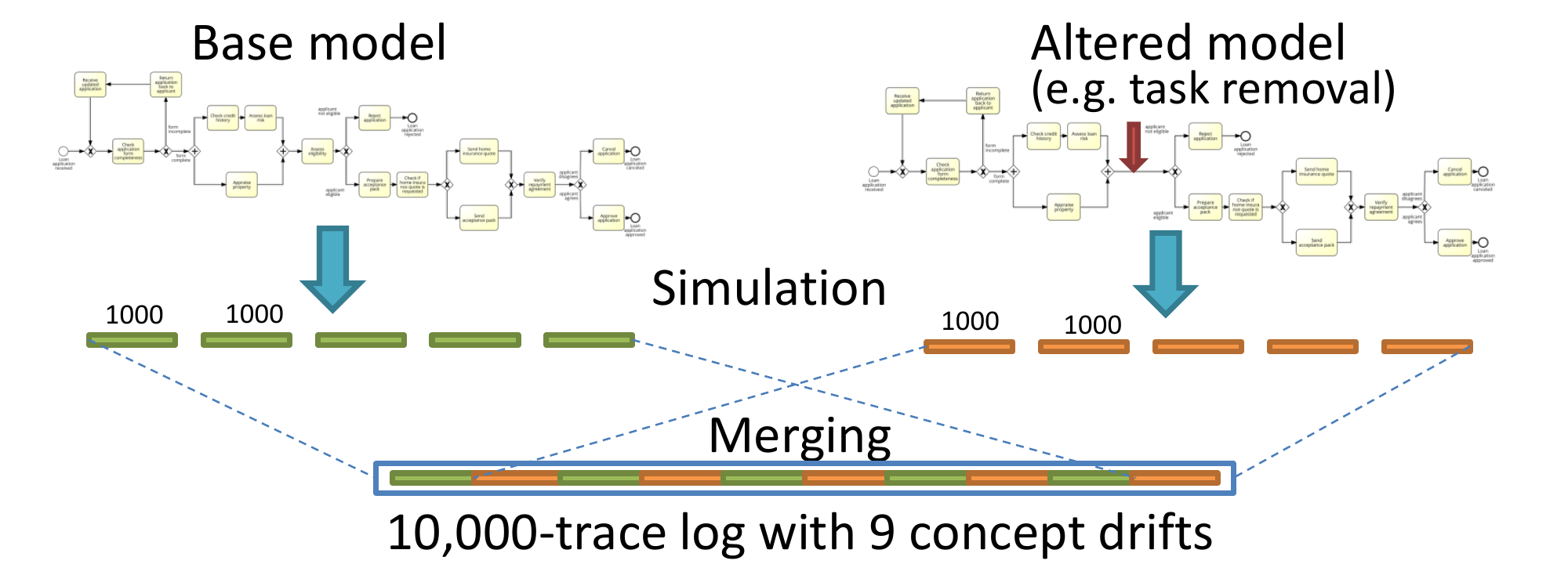} 
\caption{Event log generation with embedded process drifts}
\label{fig:loggeneration}
\end{figure}

To vary the distance between sudden drifts in the log, we generated four logs of 250, 500, 750 and 1,000 traces for the ``base'' model as well as for each of the 18 ``altered'' models, using the BIMP simulator,\footnote{Available at \url{http://bimp.cs.ut.ee}} and combined each group of 5 base logs with each group of 5 altered logs by alternating base and altered logs, in order to obtain four logs of sizes 2,500, 5,000, 7,500 and 10,000 traces for each of the 18 change patterns, leading to a total of 72 logs.\footnote{The BPMN models used for simulation, the synthetic logs and the detailed evaluation results are available with the software distribution} Figure \ref{fig:loggeneration} depicts an application of this operation to generate a log of 5,000 traces. Each log has nine sudden drifts located at multiples of 10\% of the log size, thus with an inter-drift distance ranging from 250 to 1000 traces (1000 in the example). 

%
%

\subsection{Impact of oscillation filter size on accuracy}\label{sec:phi_sudden}
\begin{figure}[hbt!]
\centering
\includegraphics[width=.8\linewidth]{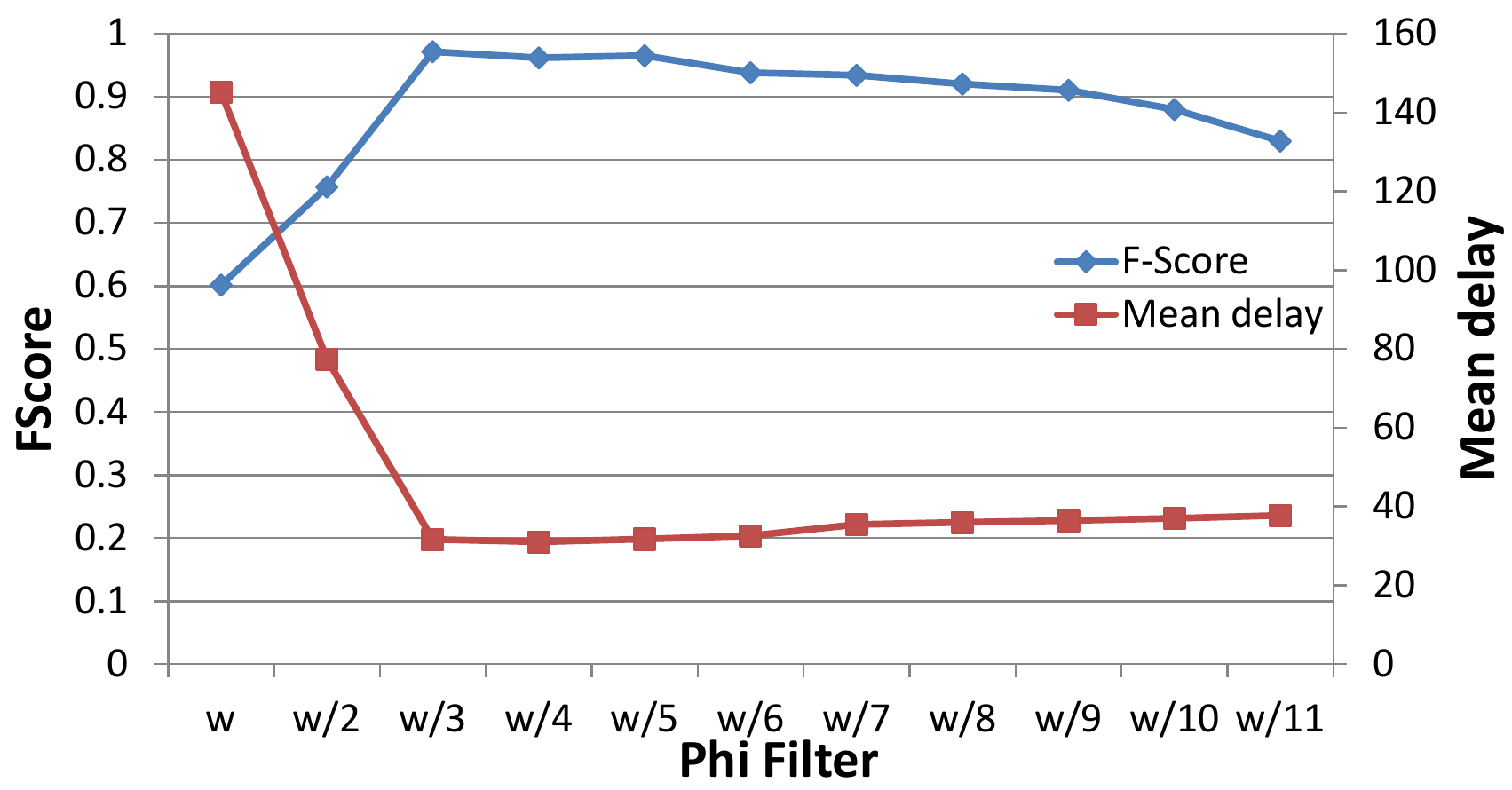} 
\vspace{-2mm}
\caption{Impact of the oscillation filter size on the accuracy of sudden drift detection}
\label{fig:phiSud}
\end{figure}

In the first experiment, we vary the oscillation filter size $\phi$ and report its impact on the average F-score and mean delay, across all 72 synthetic logs. The results are shown in Figure \ref{fig:phiSud}, where $\phi$ decreases from a size of $w$ to $w/11$ (on a nominal scale), with $w$ being the window size. 

We observe that when $\phi$ equals $w$, some sudden drifts are considered as stochastic noise (false negatives) when the P-value fails to remain under the threshold for at least $w$ consecutive statistical tests. Decreasing the size of the oscillation filter increases the sensitivity of our method which identifies more sudden drifts. This results in a higher F-score and lower mean delay, reaching the best values when $\phi$ is equal to $w/3$, with similar results until $w/5$. However, if we keep decreasing $\phi$, the method becomes less restrictive with respect to sporadic drops of the P-value, reporting false positives, which lead to a progressive decline of the F-score. In the remaining of this section, we use a $\phi$ value of $w/3$.

\vspace{-.5\baselineskip}
\subsection{Impact of window size on accuracy}
\label{sec:Fwinsize}

\begin{figure}[htb!]
	\centering
	\includegraphics[width=0.8\linewidth]{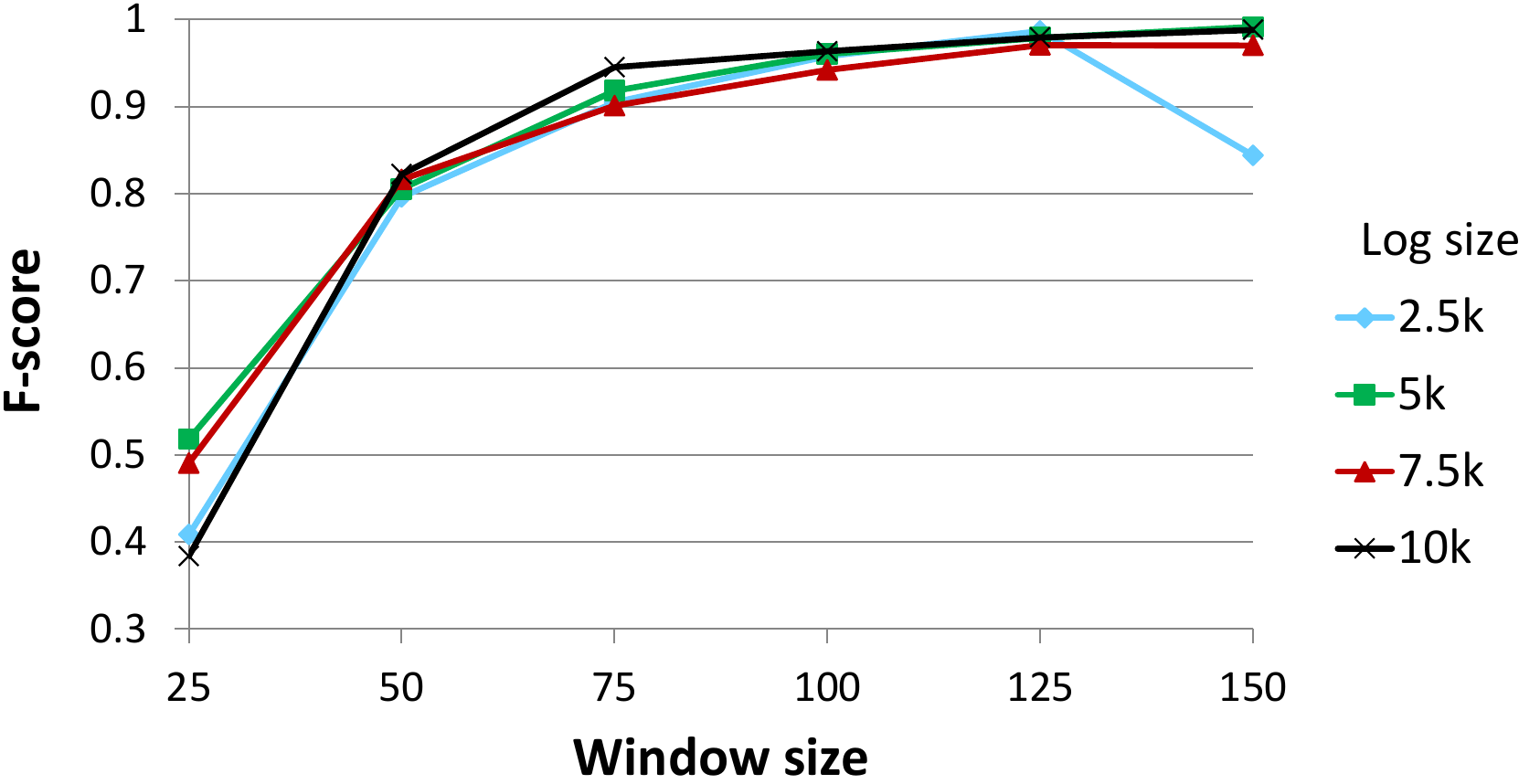}
	\vspace{-2mm}
	\caption{F-score obtained with different fixed window sizes}
	\label{fig:winsizeFscore}
\end{figure}

In the second experiment, we evaluate the impact of the window size on accuracy. For this, we executed our method with different fixed window sizes ranging from 25 to 150 traces in increments of 25, against each of the 72 logs. Figure \ref{fig:winsizeFscore} reports the F-score obtained with the four log sizes (2,500 to 10,000 traces), where for each log the F-score was averaged over the logs produced by the 18 change patterns. We also measured precision and recall separately. The corresponding graphs, reported in the online Appendix, are similar to the F-Score plotted in Figure \ref{fig:winsizeFscore} (the method errs in both false positives and false negatives).

We observe that the F-score increases as the window size grows and eventually plateaus at a window size of 150. As expected, the more data points are included in the reference and detection windows, the more likely it is for the statistical test to converge. 
This results in detecting all sudden drifts (recall of 1), with few or no false positives (precision of 0.9 or above). 
For a window size of 25 traces, the F-score is low (around 0.45). This is because the Chi-square test does not converge if more than 20\% of the data points have frequency below 5 \cite{yates}, which is often the case with a window size of 25 traces, where the distinct runs might be as low as 5-10. 
The drop in F-score at a window size of 150 for logs of 2,500 traces is not an inherent limitation of our method, but is due to having  set a drift every 10\% of the log, which equates to 250 traces for a log of 2,500 traces. Given that with a window size of 150 traces reference and detection windows aggregate 300 traces, in some cases two drifts are included within this set of traces. Hence, the method treats the two drifts as one, leading to a lower recall.


\begin{figure}[htb!]
	\centering
	\includegraphics[width=0.8\linewidth]{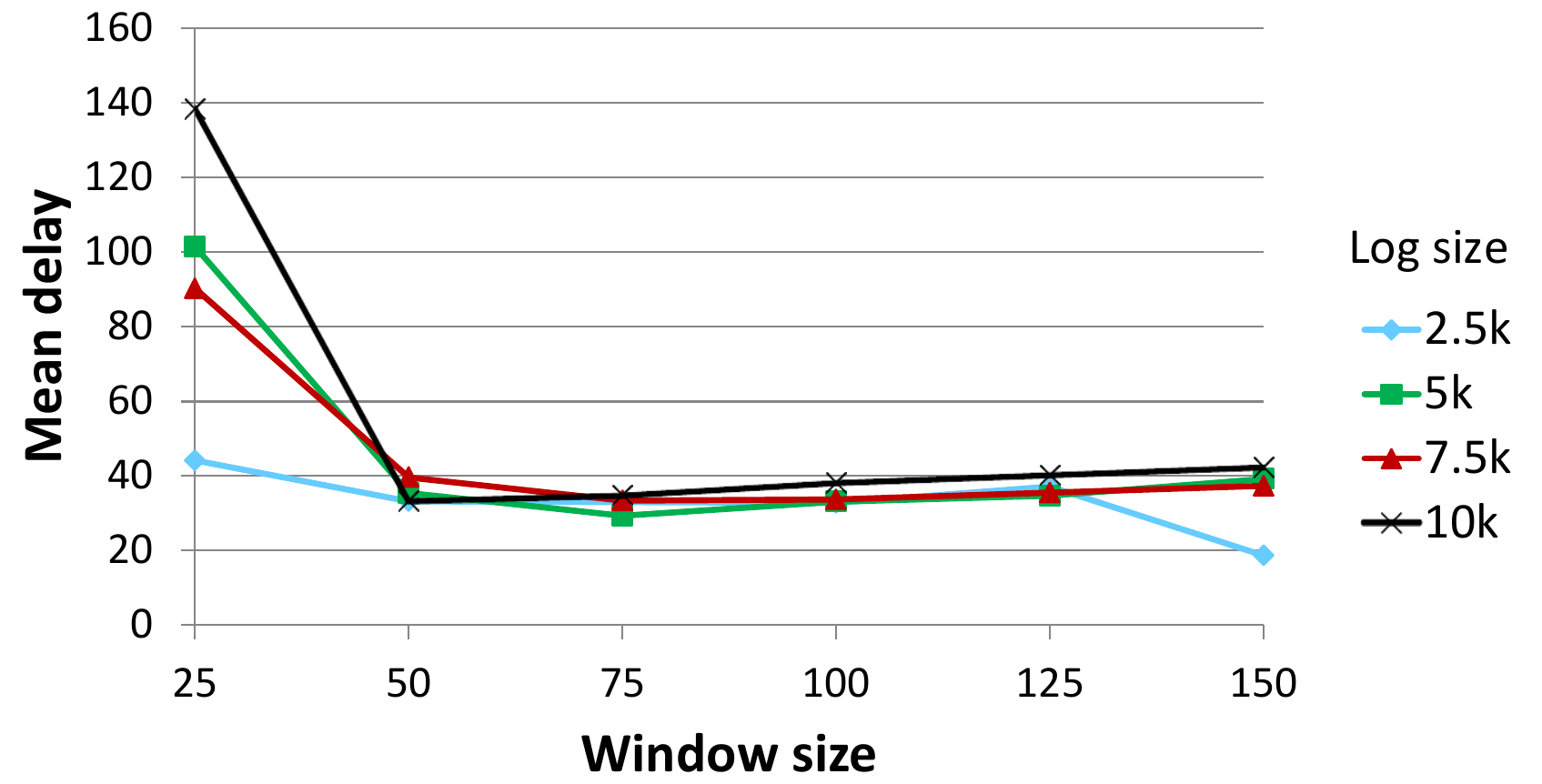}
	\vspace{-2mm}
	\caption{Mean delay obtained with different fixed window sizes}
	\label{fig:winsizeMdelay}
\end{figure}




Figure \ref{fig:winsizeMdelay} plots the mean delay for different window sizes, where the mean delay is averaged over the logs produced by the 18 change patterns, according to the four log sizes. Interestingly, after an initial high mean delay, due to the unreliability of the statistical test with low numbers of data points, the mean delay grows very slowly as the window size increases. This shows that the method is very resilient in terms of mean delay to increases in windows size, having a relatively low delay of around 40 traces when the window size is 50 or above. Similar to the results for F-score, we observe a drop in the mean delay at a window size of 150, for logs of 2,500 traces. This positive effect is due to the second drift in the composite window of 300 traces being discovered before the drift happened, with respect to the gold standard.


In summary, our method for sudden drift detection achieves high levels of accuracy both in terms of F-score (above 0.9) and mean delay (below 40 traces) in the presence of different types of sudden drifts and for different log sizes. This happens when employing a fixed window size that is at least 75 traces long, with the best trade off between F-score and mean delay being achieved with windows of 100 traces. 

We also conducted the same experiments using the trace-based representation of logs (instead of the run-based one). We observed that the obtained accuracy with the trace-based representation was consistently lower than the one with runs. This observation confirms the intuition discussed in Section~\ref{sec:method}.

\vspace{-.5\baselineskip}
\subsection{Impact of adaptive window on accuracy}

Next, we assess the impact of using an adaptive window on F-score and mean delay. For this, we compare the results obtained with the fixed window size shown in Figures~\ref{fig:winsizeFscore} and \ref{fig:winsizeMdelay}, averaged over the three log sizes of 5,000, 7,500 and 10,000 traces, with the results obtained using an adaptive window. For example, we compare the results obtained with a fixed window size of 25, with those obtained with an adaptive window initialized to 25 traces. We did not use the log size of 2,500 traces to avoid the effects of the interplay between window size and number of sudden drifts observed in logs of this size in the previous tests. 

\begin{figure}[htb!]
	\centering 
	\includegraphics[width=0.8\linewidth]{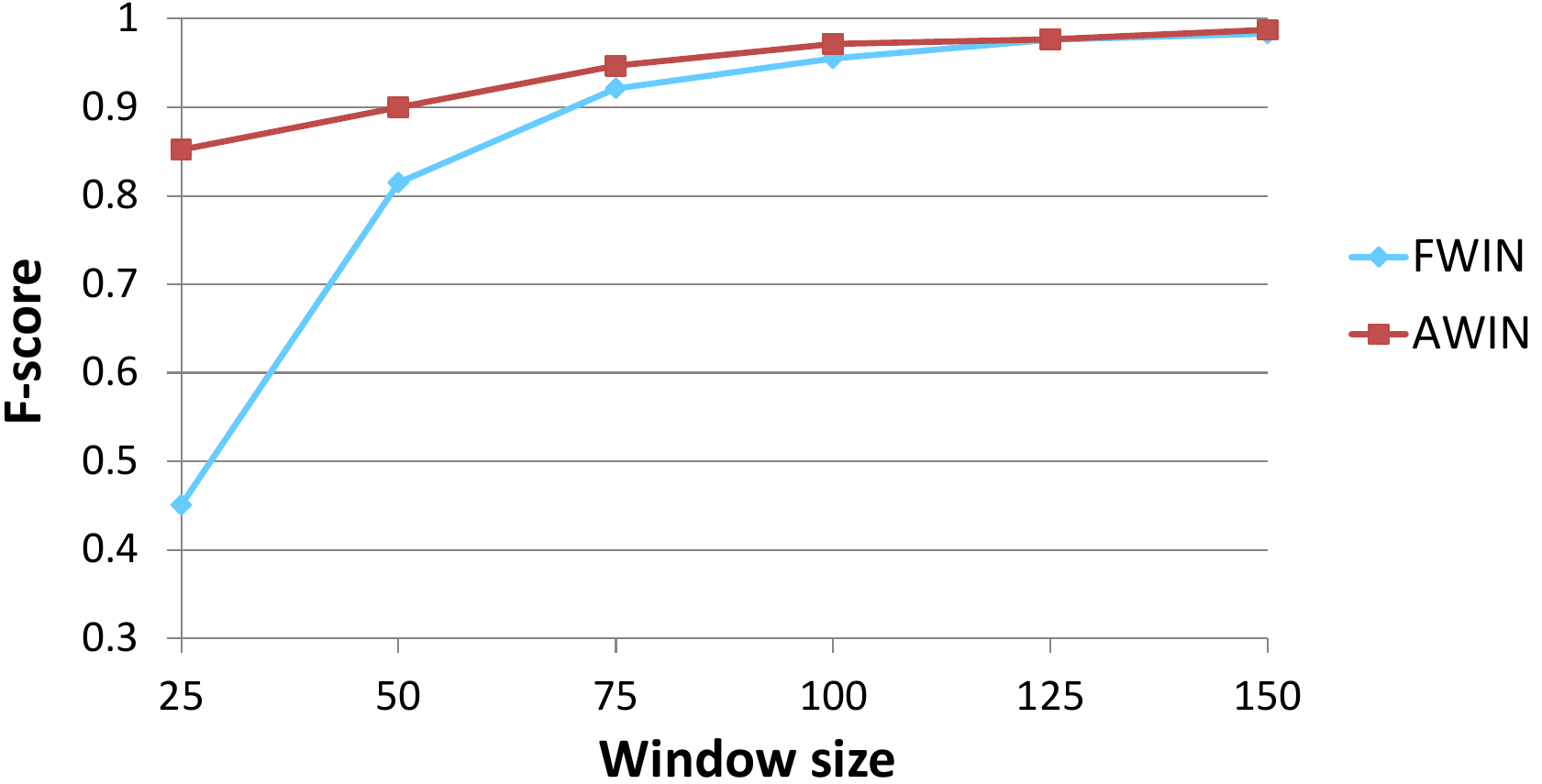}
	\vspace{-2mm}
	\caption{F-score obtained with different fixed window sizes (FWIN) vs. adaptive window sizes (AWIN)}\label{fig:winsizeAFscore}
\end{figure}

\begin{figure}[htb!]
	\centering 
	\includegraphics[width=0.8\linewidth]{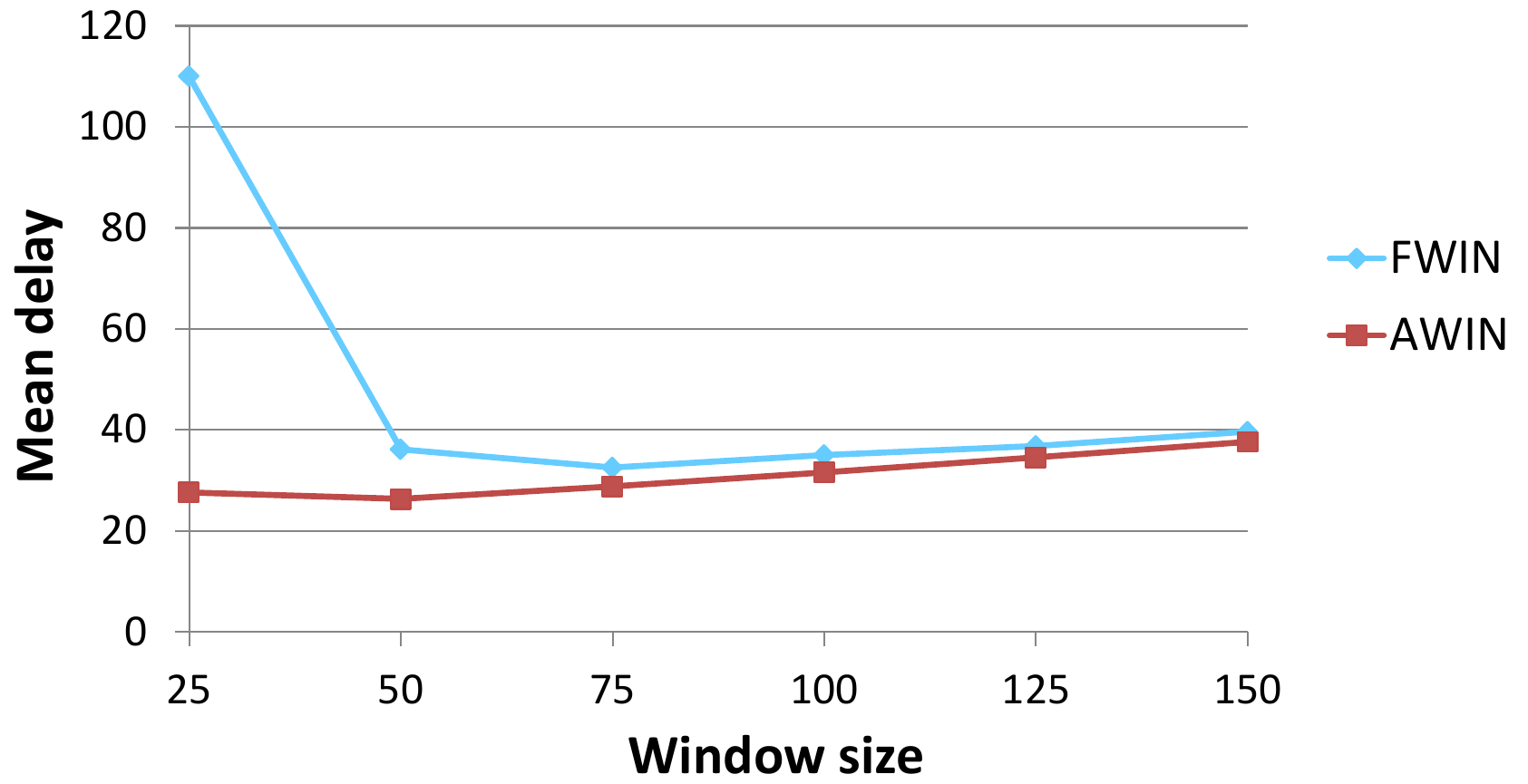}
	\vspace{-2mm}
	\caption{Mean delay obtained with different fixed window sizes (FWIN) vs. adaptive window sizes (AWIN)}\label{fig:winsizeAMdelay}
\end{figure}

Figures~\ref{fig:winsizeAFscore} and \ref{fig:winsizeAMdelay} report the results of this comparison for F-score and mean delay respectively. The use of an adaptive window outperforms the use of a fixed window both in terms of F-score and mean delay. Indeed, the ability to dynamically change the window size based on the variation observed in the log (measured as the ratio between number of distinct runs and total number of runs in the combined window), allows us to obtain an adequate number of runs (not too small, not too large) in the reference and detection windows to perform the statistical test. This leads to a higher F-score, since more data points are automatically added to the window when the variation is high. At the same time, it leads to a lower mean delay as the window size is shrank when the variation is low, since in these cases a low number of runs is sufficient to perform the statistical test. As an advantage, the adaptive window method overcomes the low accuracy (both in terms of F-score and mean delay) obtained when fixing the window size to values as low as 25 traces (F-score of 0.85 instead of 0.45, and mean delay of 28 instead of 110). This enables the method to be employed in those scenarios where the distance between drifts in the log is expected to be very low (i.e.\ in the presence of very frequent drifts) and thus keeping the mean delay as low as possible becomes essential to identify as many drifts as possible. 


\vspace{-.5\baselineskip}
\subsection{Accuracy per change pattern}\label{sec:accuracy_sudden}

As a further test on accuracy, we evaluate the relative levels of F-score and mean delay for each of the twelve simple change patterns and the six composite change patterns. For this, we fixed the window size to 100 traces, which proved to provide the best trade off in terms of F-score and mean delay, and averaged the results obtained with the fixed window, and with the adaptive window initialized to 100 traces, over the three log sizes of 5,000, 7,500 and 10,000 traces. 

\begin{figure}[htb!]
	\centering 
	\includegraphics[width=0.85\linewidth]{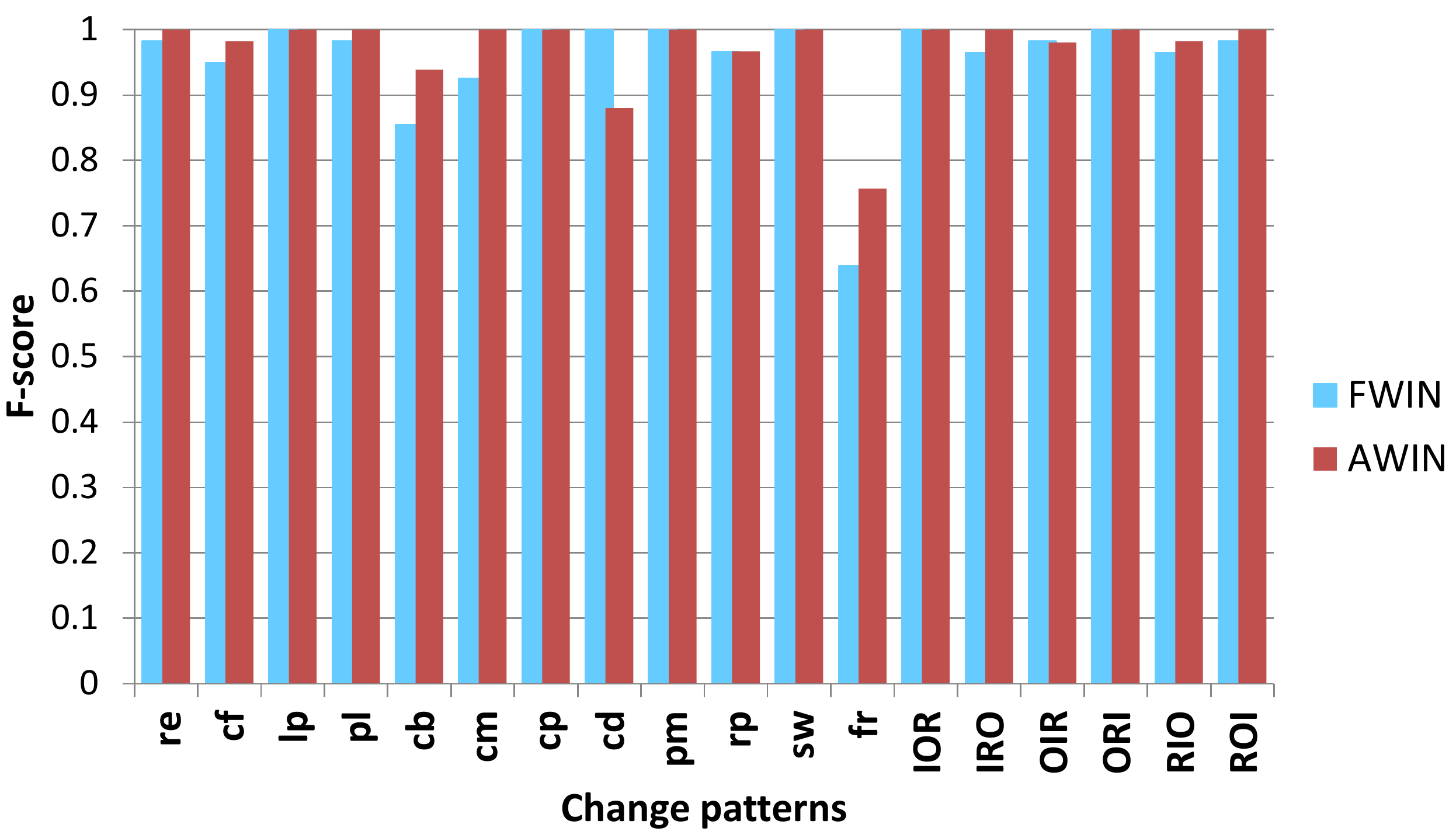}
	\vspace{-2mm}
	\caption{F-score per change pattern, obtained with fixed window size of 100 (FWIN) vs. adaptive window size (AWIN)}\label{fig:patternsFscore}
\end{figure}

\begin{figure}[htb!]
	\centering 
	\includegraphics[width=0.85\linewidth]{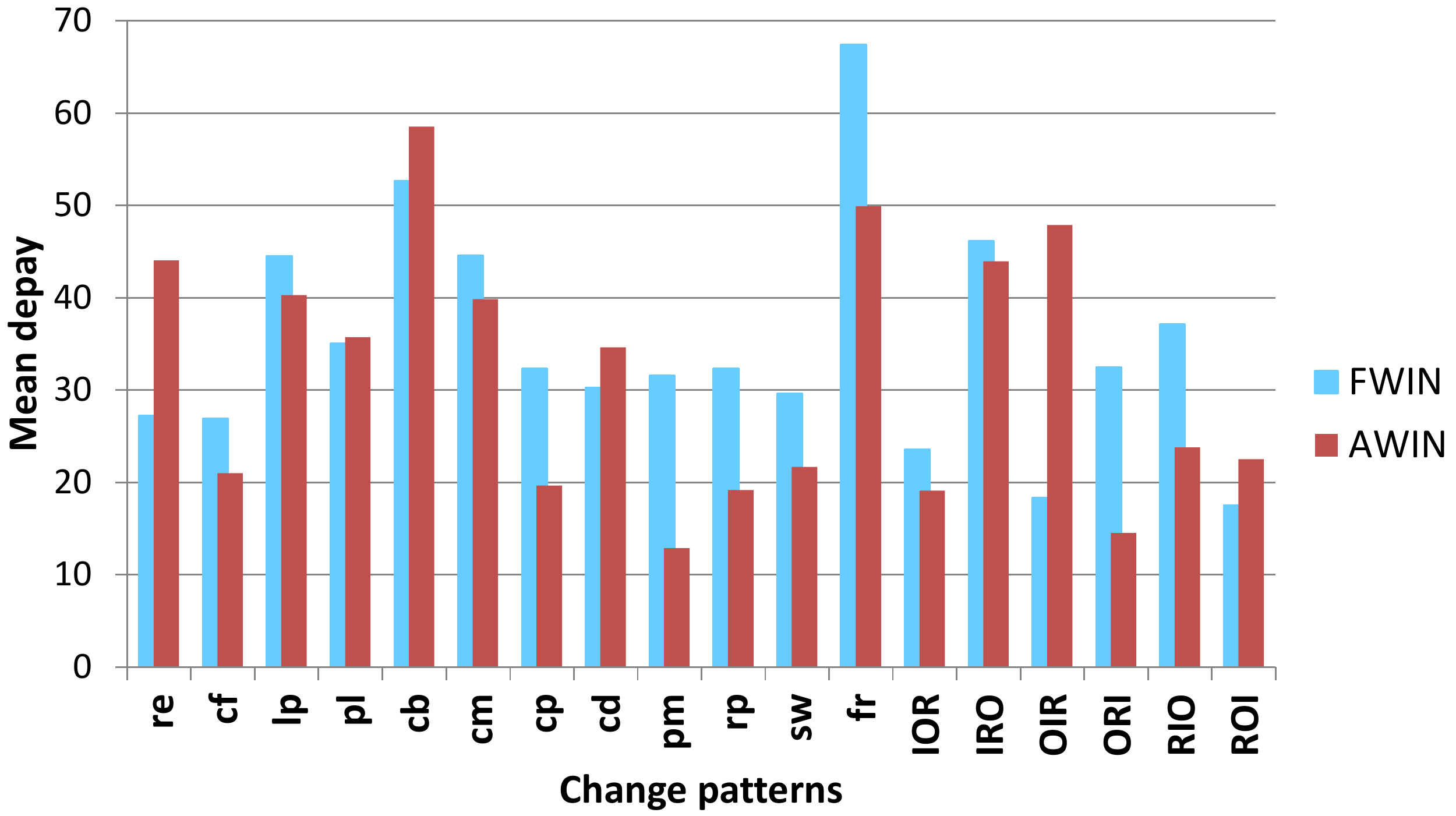}
	\vspace{-2mm}
	\caption{Mean delay per change pattern, obtained with fixed window size of 100 (FWIN) vs. adaptive window size (AWIN)}\label{fig:patternsMdelay}
\end{figure}

Figures~\ref{fig:patternsFscore} and \ref{fig:patternsMdelay} shows the results. From these we can draw the following observations. First, the use of an adaptive window enhances F-score and mean delay for the majority of patterns (16 out of 18 for F-score and 12 out of 18 for mean delay), with the F-score often being 1. Second, the method experiences a sensibly lower F-score both for fixed and adaptive windows for the \emph{frequency change} pattern (``fr''). This pattern modifies the frequency of certain event relations in the log. The low F-score is due to a low precision (lots of false positives). This is because our method is sensitive to frequency changes caused by the stochastic interference present in an event log. For example, even if the probabilities of taking two alternative branches in a process are observed to be 50\% each in the entire log, when looking at an individual window, which is a small extract of the log, these probabilities are likely to be slightly different (e.g.\ they could be 40\%-60\% instead of 50\%-50\%). This interference tricks the detection of a frequency-based drift, but can be resolved by choosing a larger window size. For example, using a fixed window of 200 traces, we obtain an F-score of 0.98 (1 if using the adaptive window) for the ``fr'' pattern.

\vspace{-.5\baselineskip}
\subsection{Comparison with baseline}
Lastly, we compare our method for sudden drift detection using an adaptive window, with the method of Bose et al. \cite{bose2014dealing}, since this is the most mature method for process sudden drift detection available at the time of writing. For this experiment we use the synthetic logs previously generated for each of the 18 change patterns, set the window size to 100 and average the results over the three different log sizes of 5,000, 7,500 and 10,000 traces.

As discussed in Section~\ref{sec:background}, the method in \cite{bose2014dealing} requires to manually select the order relations between event labels to be used as features to build the feature space which in turn is required to detect the sudden drifts. Thus, knowing the specific changes made in the altered models, we manually selected the most appropriate features for each log. 

\begin{figure}[htb!]
	\centering
	\includegraphics[width=0.85\linewidth]{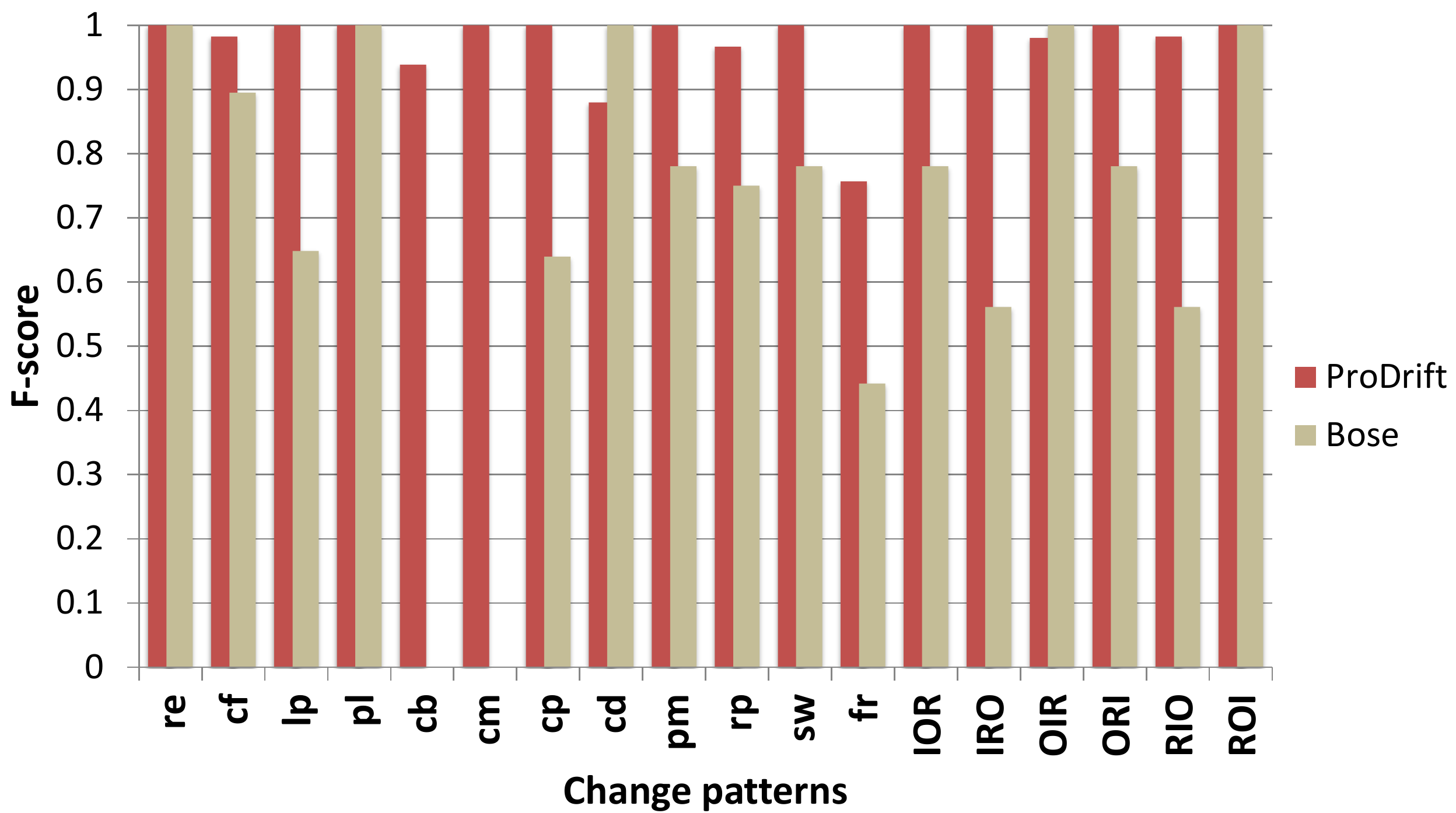}
	\vspace{-2mm}
	\caption{F-score per change pattern: our adaptive window method (ProDrift) vs. \cite{bose2014dealing} with fixed window size of 100 (BOSE)}\label{fig:boseFscore}
\end{figure}

\begin{figure}[htb!]
	\vspace*{-3mm}
	\centering
	\includegraphics[width=0.85\linewidth]{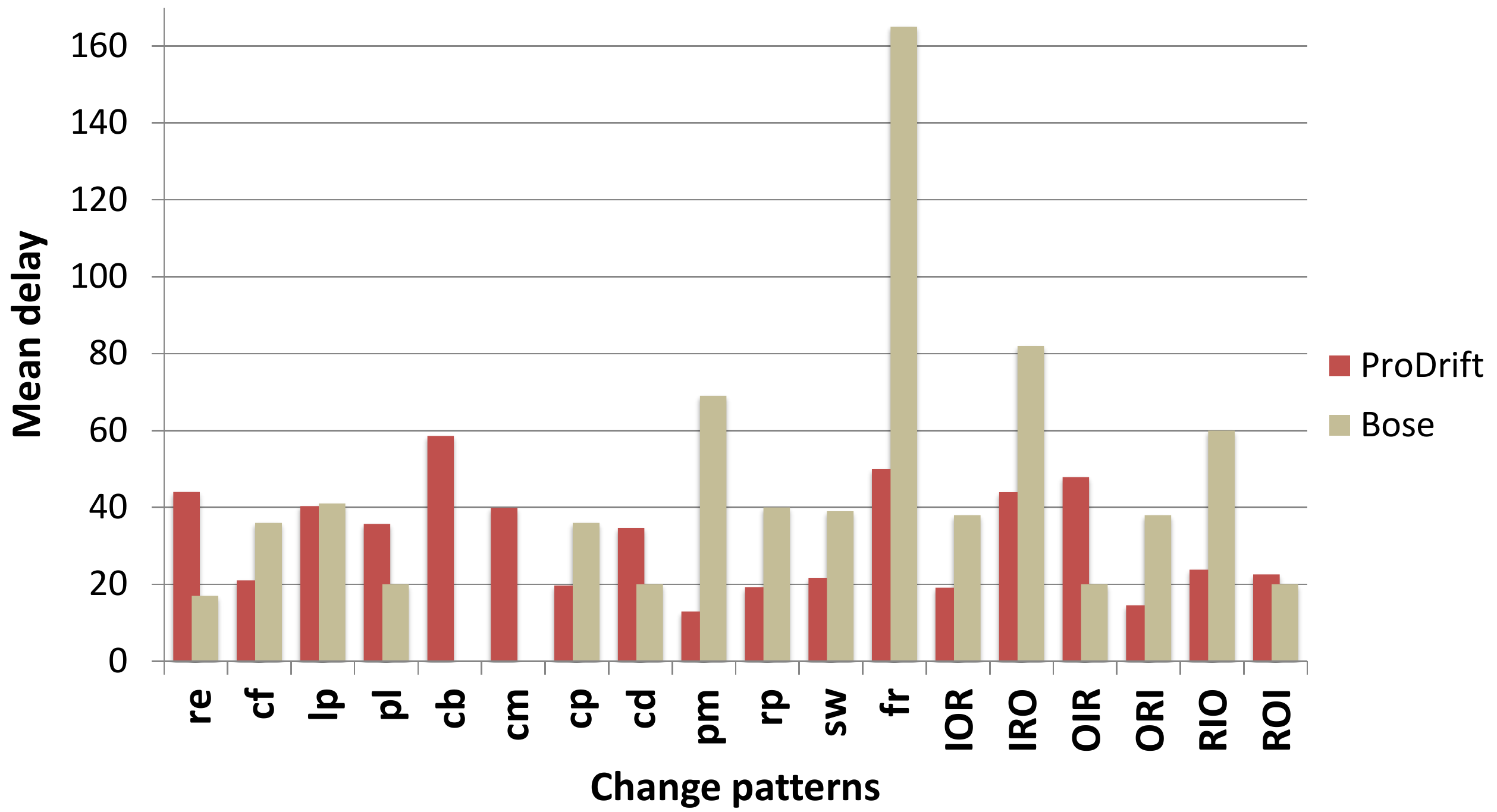}
	\vspace{-2mm}
	\caption{Mean delay per change pattern: our adaptive window method (ProDrift) vs. \cite{bose2014dealing} with fixed window size of 100 (BOSE)}\label{fig:boseMdelay}
\end{figure}

Figure~\ref{fig:boseFscore} and \ref{fig:boseMdelay} show the results of the comparison. 
Our method outperforms the method in \cite{bose2014dealing} both in terms of F-score and mean delay, achieving substantial F-score differences for ten change patterns, including ``lp'' (make fragment loopable/non-loopable), ``cp'' (duplicate fragment), ``pm'' (move fragment into/out of parallel branch) and composite patterns such as ``IOR'' and ``RIO''. This is due to the large number of false positives identified by the baseline. Further, this latter method fails to identify sudden drifts of the following changes: ``cb'' (make fragment skippable/not skippable) and ``cm'' (move fragment into/out of conditional branch), even if appropriate features are chosen.

As a final test, for each log, we selected all features in order to simulate a fully-automated application of the baseline. However, in this case the method fails to identify any drift due to a high level of false negatives, and construction of the feature space becomes an expensive task (over 15 minutes with window size of 100 traces). 

\vspace{-.5\baselineskip}
\subsection{Time performance}\label{sec:evalsuddentime}
We conducted all tests on an Intel i7 2.20GHz with 16GB RAM (64 bit), running Windows 7 and JVM 8 with standard heap space of 512MB. The time required to update the alpha-relationships, extract the runs, and perform the Chi-square test, ranges from a minimum of 0.26 milliseconds to a maximum of 2.3 milliseconds with an average of 0.5 milliseconds. These results show that the method is scalable. 

\section{Evaluation of gradual drift detection on synthetic logs}\label{sec:evaluationGrad}

In this section we discuss the evaluation of our method for detecting gradual drifts using synthetic logs. First, we describe the dataset generation and the specific evaluation criteria we employed to assess the accuracy of this method. As we will see, these differ from those used for assessing the accuracy of the method for detecting sudden drifts. Once we have defined how to measure accuracy, we study the impact of the oscillation filter size on accuracy, then evaluate the accuracy per change pattern and compare the results with a baseline approach for gradual drift detection. Finally, we report on time performance. 

We did not study the impact of window size nor that of using an adaptive window on accuracy, because the gradual drift detection method post-processes the sudden drifts already detected with the basic method. Hence, we carried out these experiments using the best settings obtained in the previous experiments, i.e.\ using an adaptive window initialized to 100 traces.

\vspace{-.5\baselineskip}
\subsection{Dataset generation}
We generated a dataset of 18 logs, one for each change pattern, by following the same procedure as in Section~\ref{sec:patterns}, i.e.\ we combined the base log with one of the 18 altered logs (resulting from the injection of 12 simple patterns and 6 composite patterns) in an interleaving manner. 


However, in order to simulate gradual rather than sudden changes, we combine the base log with an altered log in a different manner. We start by sampling traces from the base log only. As the number of traces increases, we reduce the probability of sampling from the first log while increasing the probability of sampling from the altered log, 
until we only sample traces from the altered log. This operation, known as probabilistic gradual drift \cite{Minku2010}, results in gradually decreasing the proportion of traces from the first log and increasing the proportion of traces from the second log in the gradual drift interval
. We repeat this operation by inverting the order of the base and altered log, so as to introduce and later remove the same change, for a total of 10,000 traces per log.


To sample the behavior from the two logs, we use a linear probability function as in \cite{Minku2010}, with a slope of 0.2\%. In other terms, from a probability starting at 1 (resp.\ 0) for the first (resp.\ second) log, the probability of a trace to be selected from the first log (resp.\ the second log) decreases (resp.\ increases) by 0.002 every time a new trace is sampled, to reach 0 (resp.\ 1) after 500 traces. This leads to a gradual drift interval of 500 traces. Moreover, having the base and altered logs of 1,000 traces each, each gradual drift interval includes 25\% of the behavior of each log, so as to have a significant portion of that behavior in each gradual drift. In these settings, we choose an inter-drift distance of 500 traces, to avoid any inference between two consecutive gradual drifts. This led to the 18 logs being 10,000 traces each, with the first drift starting at trace number 751, with a total of nine drifts per log.


\vspace{-.5\baselineskip}
\subsection{Criteria for evaluating accuracy}
To assess the accuracy of our gradual drift detection method we also use F-Score and mean delay, but defined in a slightly different way. To compute precision and recall for the F-Score, we say that a detected drift is a \emph{true positive} if its detected interval includes the central point of the interval of the actual gradual drift, i.e.\ the point that is halfway from each end of the actual gradual drift interval, otherwise we consider it as a \emph{false positive}. For instance, if the actual gradual drift happened between trace numbers 751 and 1,250, the central point would be at trace number 1,000. In this case, a gradual drift that it detected with any interval that includes the trace number 1,000, e.g.\ [800-1,300], is considered as a true positive. If the detected drift interval does not include the trace number 1,000, e.g.\ [1,200-1,500], the gradual drift is treated as a false negative.

\begin{figure}
  \centering
	\includegraphics[width=.9\linewidth]{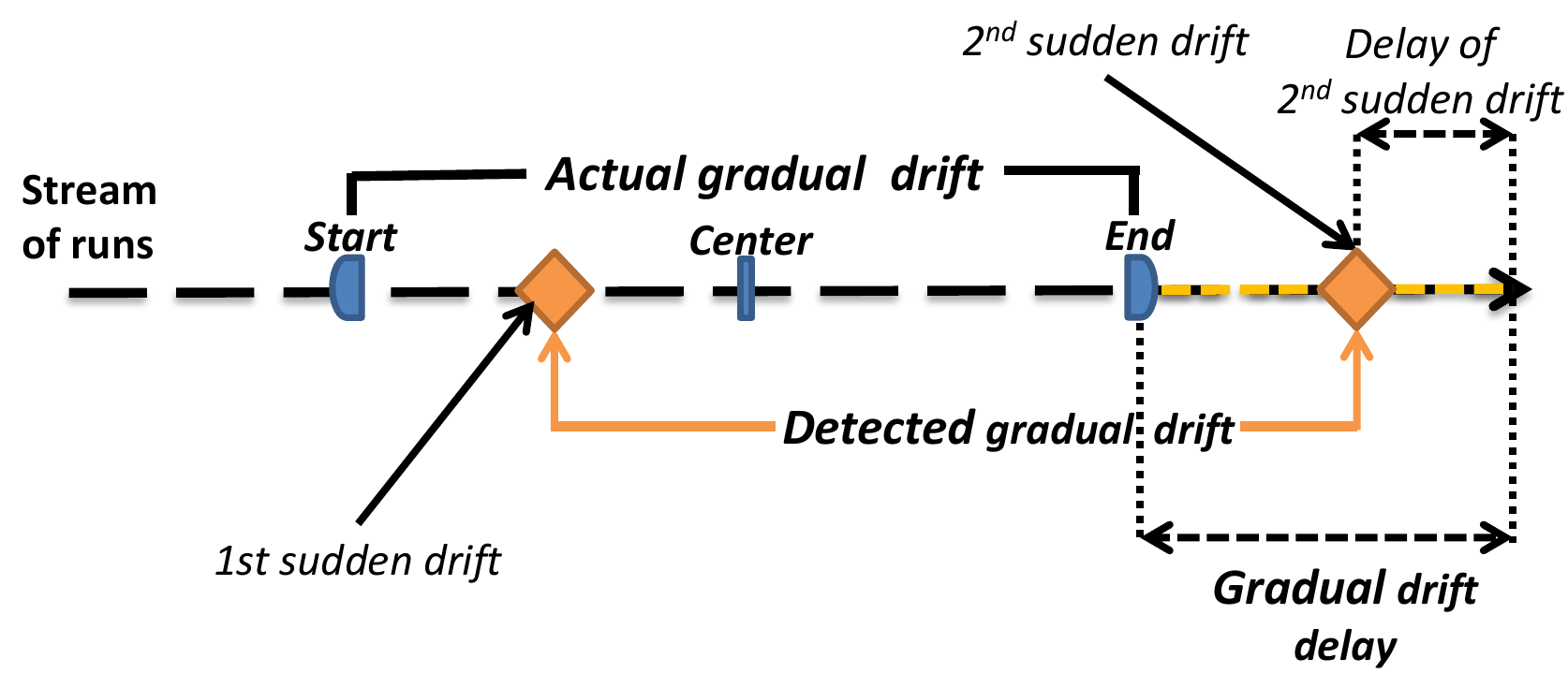} 
	\vspace{-2mm}
	\caption{A true positive gradual drift and its mean delay}
	\label{fig:gradcriteria}
	\vspace{-\baselineskip}
\end{figure}

The notion of a mean delay for gradual drifts is not defined in the literature. 
Using the same intuition behind the delay of a sudden drift, we therefore define the delay of a gradual drift as the number of traces needed to detect the end point of the actual gradual drift interval. This is equal to the delay required to identify the second sudden drift plus the distance between this point and the actual end point of the gradual drift, as shown in Figure \ref{fig:gradcriteria}. Accordingly, the mean delay for detecting gradual drifts is the arithmetic mean of all the calculated delays in each log.


\vspace{-.5\baselineskip}
\subsection{Impact of oscillation filter size on accuracy}\label{sec:phi_gradual}

\begin{figure}[htbp!]
\centering
\includegraphics[width=.85\linewidth]{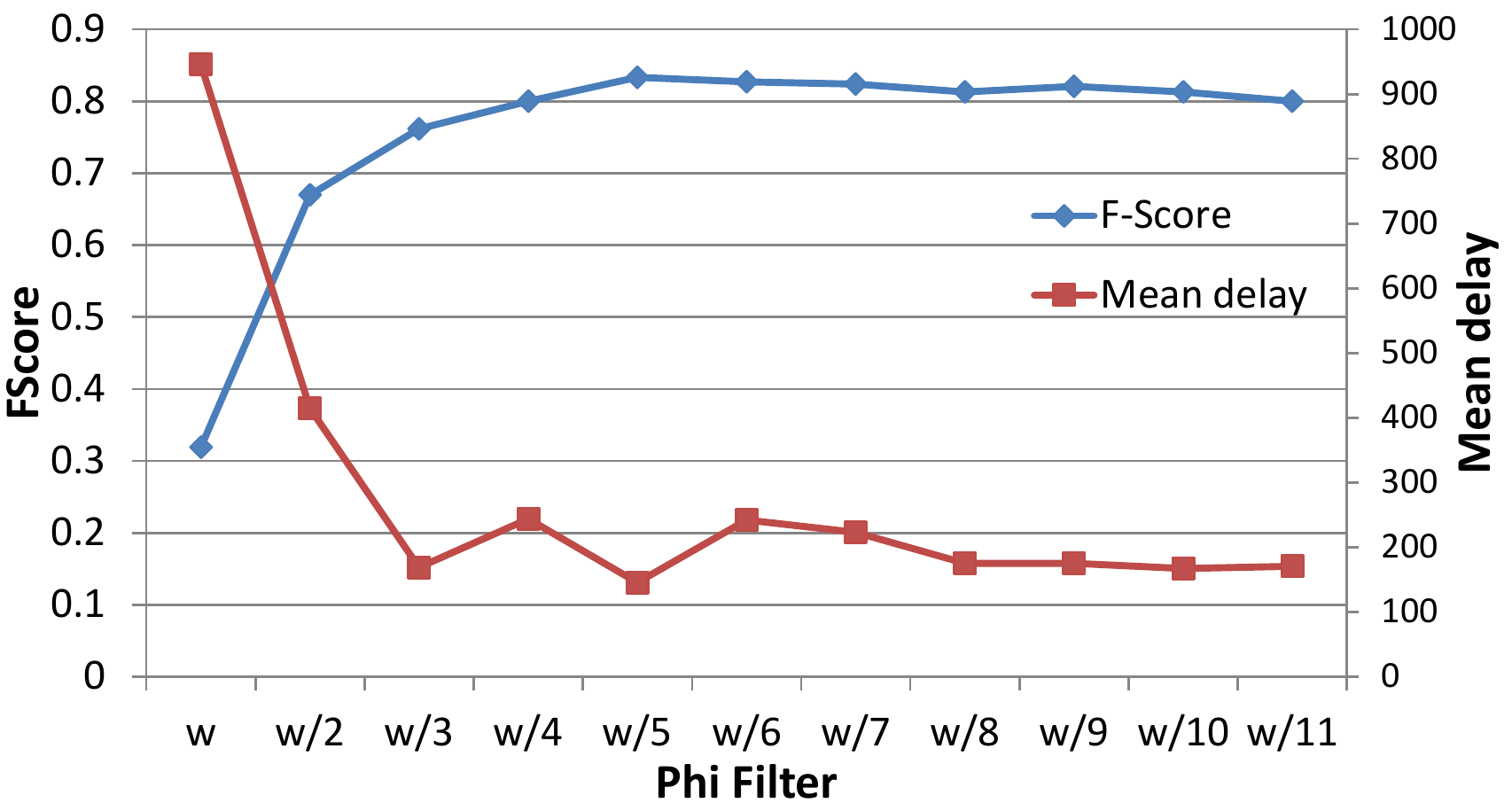} 
\vspace{-2mm}
\caption{Impact of the oscillation filter size on the accuracy of gradual drift detection}
\label{fig:gradfilter}
\end{figure}

Similar to Section~\ref{sec:phi_sudden}, we first measure the impact of the size of the oscillation filter $\phi$ on F-score and mean delay. The results are reported in Figure \ref{fig:gradfilter}. 
We observe that a $\phi$ equaling to $w$ leads to very poor results, due to the fact that some drifts are considered as stochastic noise when the P-value fails to remain under the threshold for more than $w$ consecutive statistical test, coupled with the fact that missing out one sudden drift leads to missing out the whole gradual drift. Moreover, due to the way true positives are defined, some gradual drifts may be discarded because these are detected too late, i.e.\ their interval does not include the central point of the actual gradual drift interval.

Decreasing the size of $\phi$ increases the sensitivity of our method which misses out less sudden drifts. This leads to a higher F-score and lower mean delay, with the best values obtained at $\phi=w/5$. Hence, we hereon set $\phi$ to $w/5$.

Similar to the trend observed for sudden drifts, if we keep decreasing $\phi$ the method becomes less restrictive with respect to sporadic drops of the P-value, reporting an increasing number of false positives, which leads to a decline of the F-score. The oscillations on the mean delay at $w/4$, $w/6$ and $w/7$ are due to using a smaller set of logs (18) than the set used in the sudden drift experiments (72 logs).


\vspace{-.5\baselineskip}
\subsection{Accuracy per change pattern}

\begin{figure}[htb!]
	\centering 
	\includegraphics[width=0.85\linewidth]{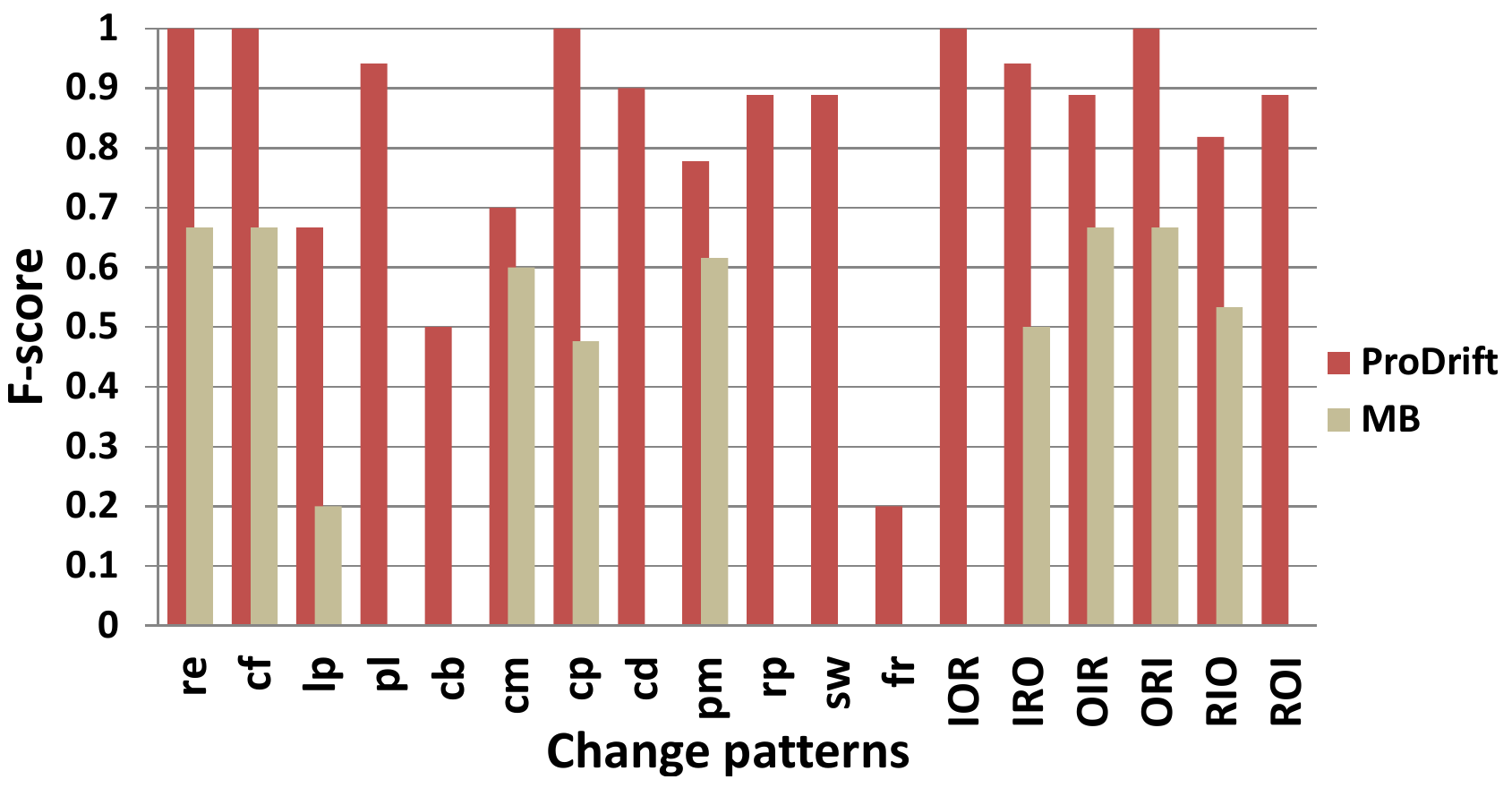}
	\vspace{-2mm}
	\caption{F-score per change pattern, obtained with our gradual drift method (\textit{ProDrift}) vs. \textit{MB} \cite{Martjushev2015} with default settings}
\label{fig:gradBaselineFscore}
\vspace{-1\baselineskip}
\end{figure}
\begin{figure}[htb!]
	\centering 
	\includegraphics[width=0.9\linewidth]{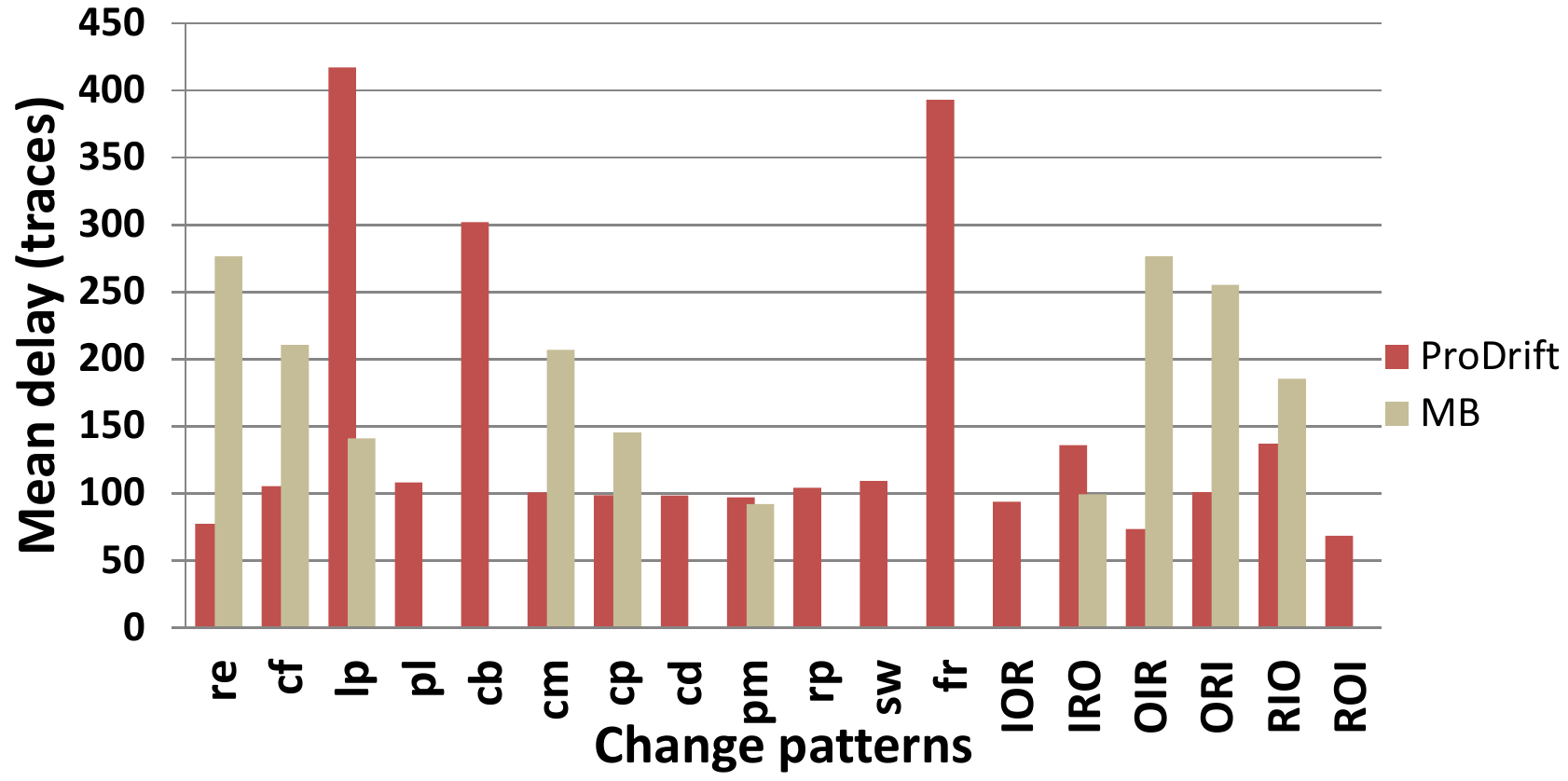}
	\vspace{-2mm}
	\caption{Mean delay per change pattern, obtained with our gradual drift method (\textit{ProDrift}) vs. \textit{MB} \cite{Martjushev2015} with default settings}
\label{fig:gradBaselineMdelay}
\end{figure}

Figures \ref{fig:gradBaselineFscore} and \ref{fig:gradBaselineMdelay} report the F-score and mean delay of our method (labeled as \emph{ProDrift}) for each change pattern. The majority of change patterns can be detected with a reasonably high accuracy (F-score of about 0.8 and mean delay of about 100 traces). This means on the one hand that most of actual gradual drifts where correctly detected (recall). On the other hand, most of intervals built with the two consecutive sudden drifts where the first sudden drift is the end of a gradual drift and second sudden drift is the start of the succeeding gradual drift are correctly classified as non gradual drift intervals (precision).

However, compared to the results for sudden drift detection (cf. Section~\ref{sec:accuracy_sudden}), our method for gradual drift detection achieves an F-score lower than 0.7, and a relatively high mean delay, for three patterns: ``lp'' (make fragment loopable/non-loopable), ``cb'' (make fragment skippable/not skippable) and ``fr'' (change branching frequency). Gradual drifts are in fact less likely to be detected than sudden drifts since missing the start or the end of the gradual drift interval results in missing the gradual drift altogether. For example, as we have already pointed out, if the gap between two consecutive sudden drifts is shorter than the aggregate of the reference and the detection windows size then one of the sudden drifts might be missed out. If these two sudden drifts were actually delimiting a gradual drift then the gradual drift would be missed too.
Moreover, the sudden drifts at the start or end of a gradual drift are not as distinguishable as opposed to an abrupt sudden drift since they happen progressively. These two reasons increase the likelihood of missing any subsequent gradual drift.


In particular, the lowest accuracy is obtained with the ``fr'' pattern, which is in-line with the results for sudden drift detection. Our method has difficulties to spot all the sudden drifts for this pattern since the change in the frequencies of runs happens gradually and may be misclassified as sporadic stochastic oscillation, leading to false negatives. 
In order to verify this explanation, we reduced $\phi$ from $w/5$ to $w/10$, which led to an F-score of 0.66 (recall 0.83, precision 0.55). To a lesser extent, this same phenomenon happens with the other two patterns (``lp'' and ``cb''). For example, in the case of ``cb'', a task that is skippable (old behavior) gradually becomes non-skippable (new behavior) over an interval of 500 traces. Here the method is tricked by the fact that already in the old behavior, a skippable task is observed in some runs but not in others. 
As a verification step, we also tested our method for gradual drift detection with the 72 synthetic logs used in Section~\ref{sec:patterns}, which contain sudden drifts only. However, no gradual drifts were reported (precision~=~1). 


\vspace{-.5\baselineskip}
\subsection{Comparison with baseline}\label{subsec:gradBaseline}

We compared our method for gradual drift detection with that by Martjushev et al. \cite{Martjushev2015}, which is the only work available in the literature for detecting gradual process drifts. Unlike ours, this method requires the user to specify the type of drift to be detected. If gradual drift is selected, the user also needs to specify the minimum and maximum size of the gradual drift interval referred to as the \emph{gap} in the tool.
\footnote{We used the default settings for this method: use of adaptive window for both population size and gradual drift gap, the latter varying between 50 and 500; step size of 10; use of the KS-test and the J-measure feature over all activity pairs; P-value threshold of 0.4.}


Figures \ref{fig:gradBaselineFscore} and \ref{fig:gradBaselineMdelay} report the results of the comparison. In terms of F-Score, the baseline method cannot detect eight of the 18 change patterns. In terms of mean delay, our method requires an average of 100 traces to detect a gradual drift, except for the patterns with low F-score where the mean delay reaches 400 traces in the worst case. Yet, our method outperforms the baseline in 15 change patterns out of 18. Notably, Martjushev et al. achieve a shorter mean delay for the loop change pattern (``lp''), but only detects one of its nine injected drifts (F-Score of 0.2).

A closer look at these results showed that the baseline method tends to detects two gradual drifts at the start and end of an actual gradual drift. While one of these two drifts is considered as a true positive, the other one is treated as a false positive, which negatively impacts on accuracy. This issue is due to the gap between the reference and the detection window, which is introduced by design in order to identify gradual drifts. 

\vspace{-.5\baselineskip}
\subsection{Time performance}
We conducted the above tests on the same machine as we used for gradual drift detection (cf. Section~\ref{sec:evalsuddentime}). Using the JOptimizer solver, it took on average 0.9 milliseconds across all 18 logs to solve each inequality generated from every two consecutive sudden drifts. These results show that the method for gradual drift detection scales well to large logs. 

\section{Evaluation on real-life logs}\label{sec:evaluationReal}

As a final step, we evaluated our methods for sudden and gradual drift detection against two event logs originating from the claims management system of a large Australian insurance company. 

\vspace{-.5\baselineskip}
\subsection{Sudden drift detection}
The first log consists of 4,509 traces with 29,108 total events of which 12 are distinct events. It records executions of a claims handling process for motor insurance claims, that were performed over a period of 13 months between 2011 and 2012.

\begin{figure}[htb!]
	\centering
	\includegraphics[width=0.9\linewidth]{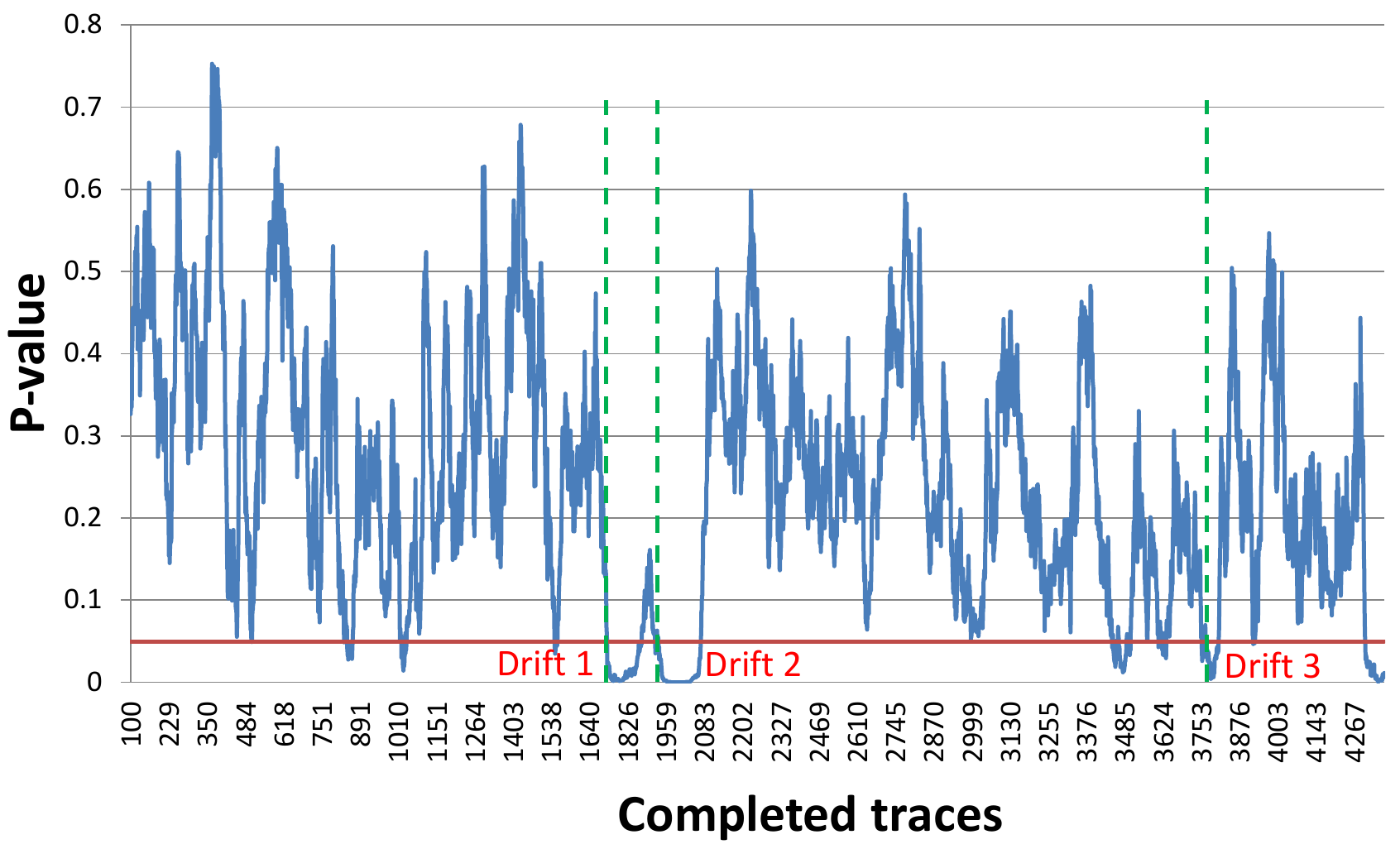} 
	\caption{Plot of the Chi-square test results}\label{fig:insurancePval}
\vspace{-\baselineskip}
\end{figure}

We initialized the adaptive window to 100 traces. The method took 4.51 seconds to check the whole log and returned three sudden drifts at 1,769, 1,911 and 3,763 traces, as shown by the results of the Chi-square test in Figure~\ref{fig:insurancePval}. In this plot we can also see stochastic oscillations that were automatically filtered out by our method. The techniques in \cite{bose2014dealing} did not report any sudden drift. 

\begin{figure}[htb!]
\centering
\includegraphics[width=.9\linewidth]{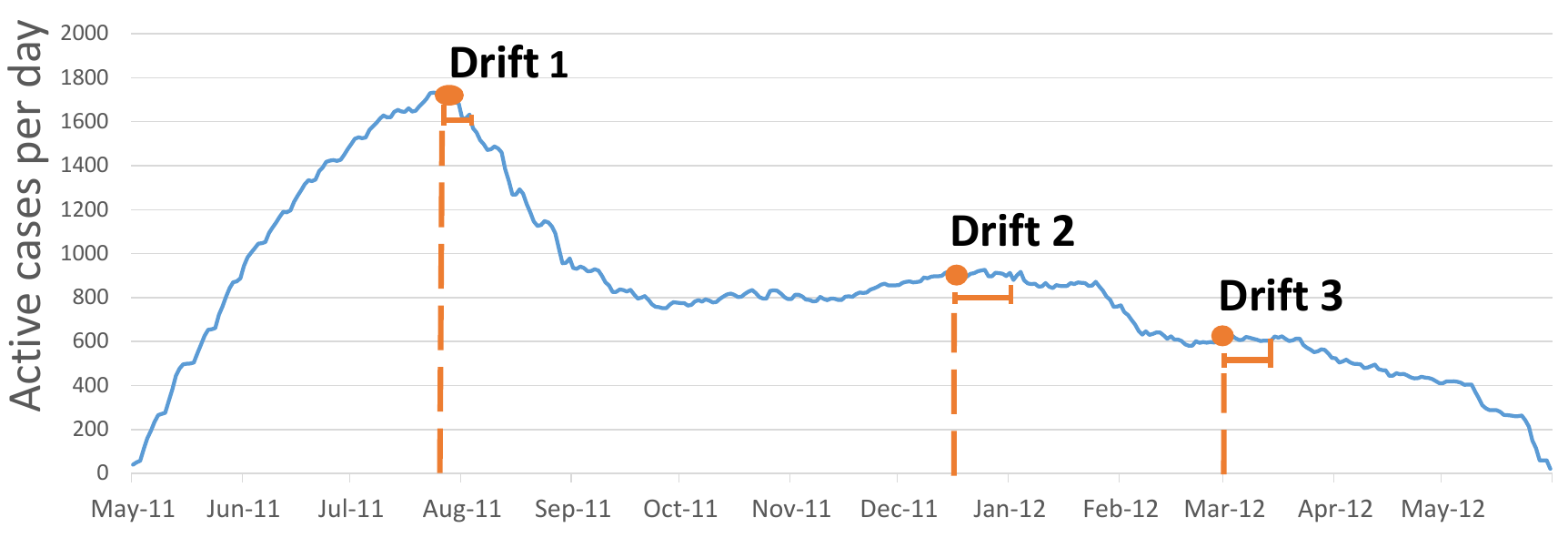} 
\vspace{-2mm}
\caption{The position and delay of the three sudden drifts detected by our method w.r.t. active cases over log timeline}
\label{fig:insuranceActivcases}
\end{figure}

We then validated the results with a business analyst from the insurance company, who confirmed that the three drifts correspond to a new major release (Drift 1) and two minor releases (Drifts 2 and 3) of the claims management system. These releases led to various changes in the claim handling process supported by the system, e.g.\ the removal of a manual task for reviewing the claim correspondence and the replacement of a manual task for checking the invoice with an automated one, with the purpose of reducing the total number of open claims. The effects of these changes are confirmed by the distribution of the number of active cases over the log timeline, shown in Figure~\ref{fig:insuranceActivcases}, which we have annotated with the position of the drifts identified by our method and the delays in reporting these drifts. We can see that each drift is associated with a drop in the number of active cases, confirming the effectiveness of the new releases on process performance.  

\begin{figure}[htb!]
	\centering 
	\includegraphics[width=0.9\linewidth]{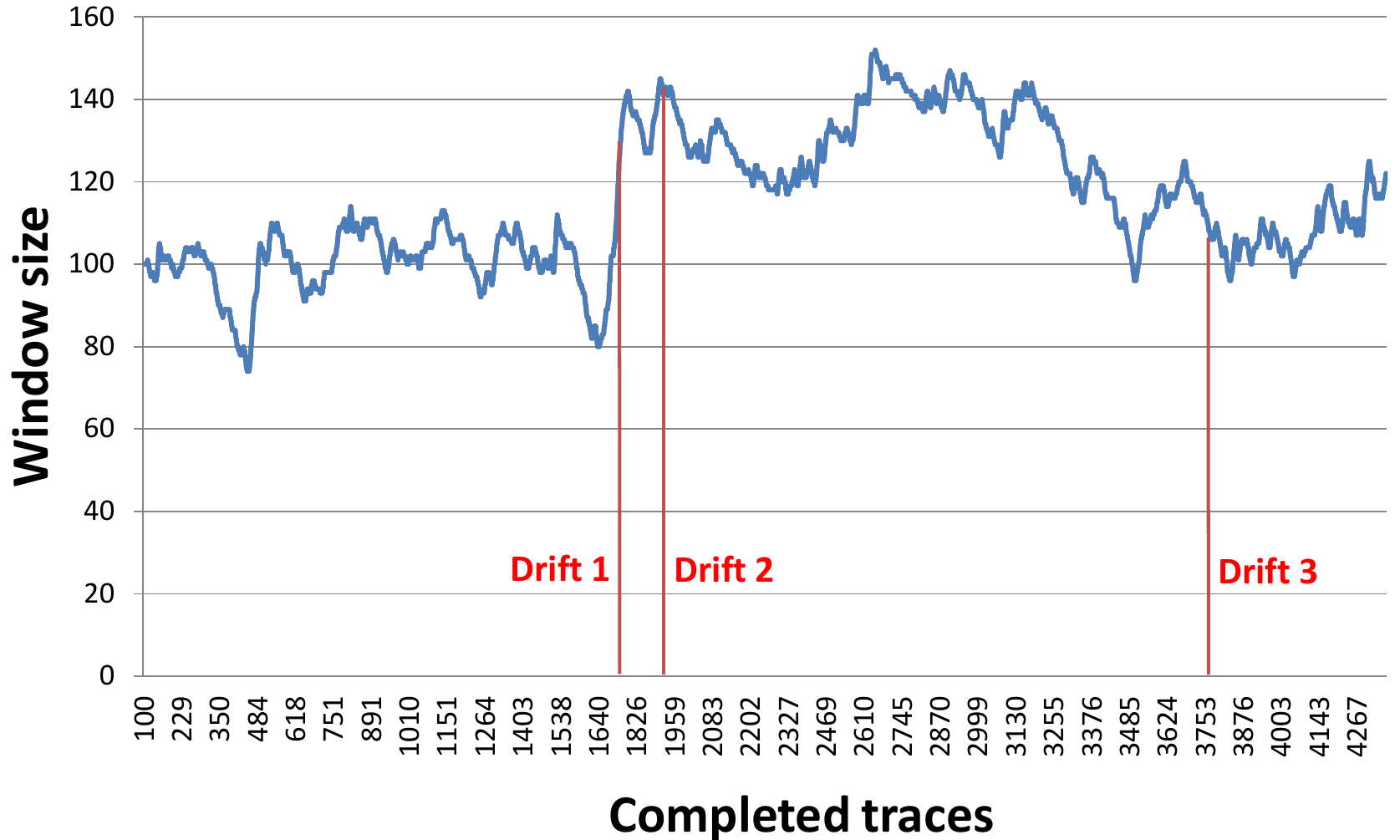}
	\vspace{-2mm}
	\caption{Plot of the adaptive window size}\label{fig:insuranceWinsize}
	\vspace{-\baselineskip}
\end{figure}

The delay in detecting the first two drifts is longer than the delay in detecting the last drift. This is due to a higher level of variation in the first part of the log (due to the more manual nature of the business process), which led our method to increase the size of the adaptive window. This is confirmed by Figure~\ref{fig:insuranceWinsize}, which shows how the window size varies according to the number of completed traces. Here we can see that the detection of Drift 1 and 2 is associated with a larger window size (131, resp.\ 143) than the size required to detect Drift 3 (size 109).

\vspace{-.5\baselineskip}
\subsection{Gradual drift detection}

\begin{figure}[htb!]
	\centering
	\includegraphics[width=0.9\linewidth]{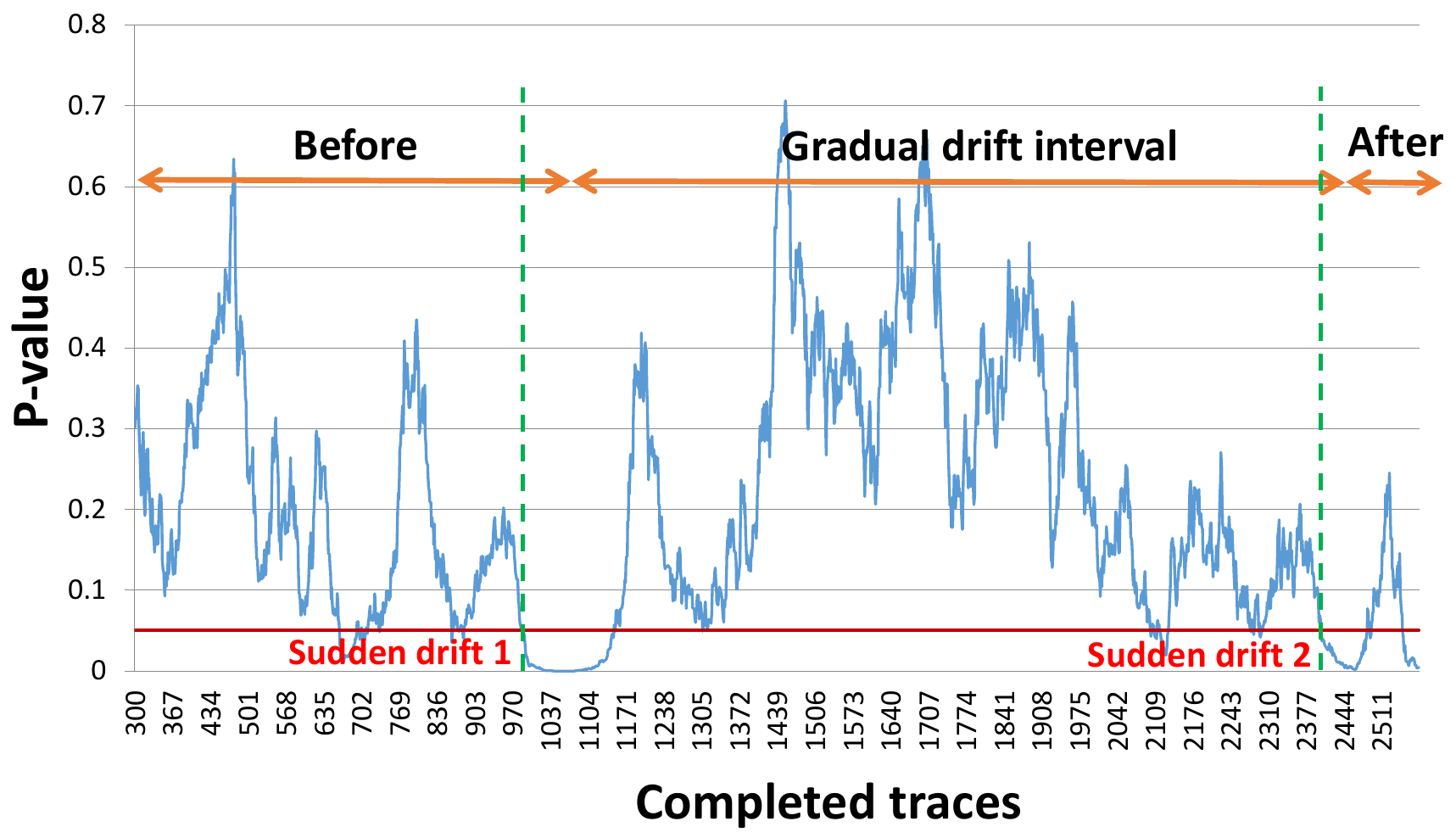} 
	\vspace{-2mm}
	\caption{Plot of the Chi-square test results}
	\label{fig:gradinsurancePval}
	\vspace{-\baselineskip}
\end{figure}

The second log also records motor insurance claims, but related to a different insurance brand. This log goes over the same timeframe as the first log, and consists of 2,577 traces and 17,474 events, of which 14 are distinct events. Our method took 1.79 seconds to complete the analysis, and reported one gradual drift between trace numbers 1145 and 2253 and no sudden drifts. Figure \ref{fig:gradinsurancePval} plots the P-value, showing the position of the two sudden drifts that delimit the interval of the detected gradual drift. We also applied the baseline method in \cite{Martjushev2015} on this log, which reported a gradual drift between trace numbers 328 and 650.



\begin{figure}[htb!]
	\centering 
	\includegraphics[width=0.9\linewidth]{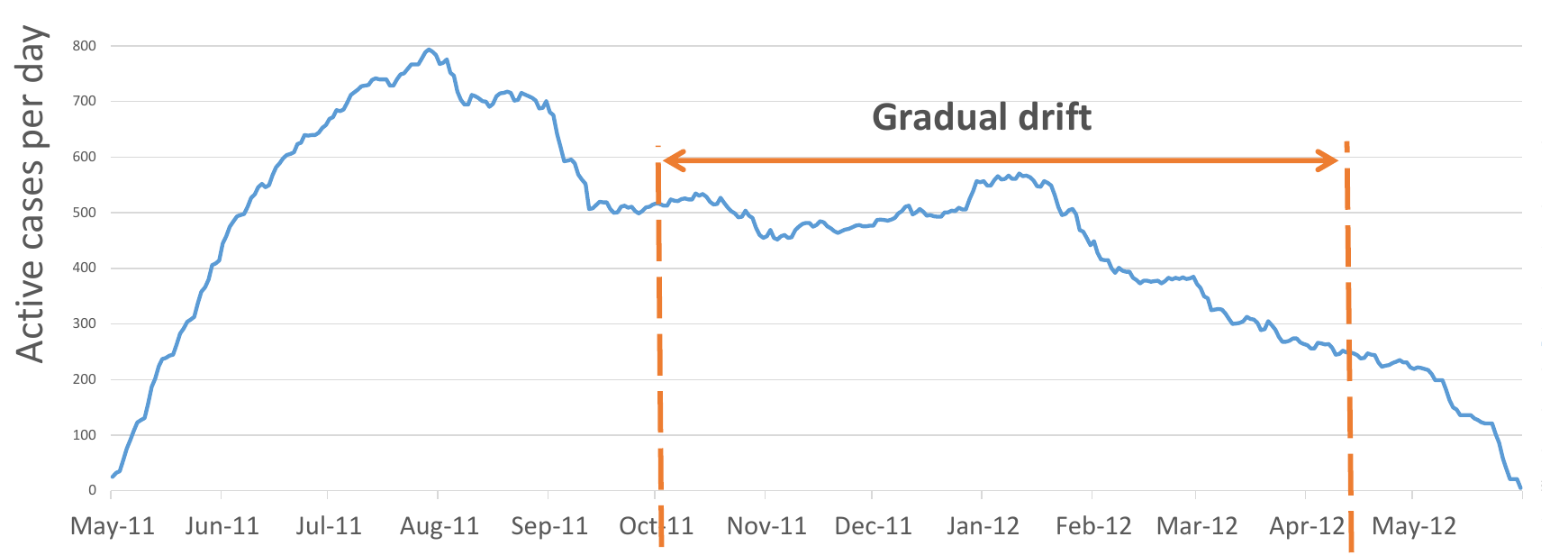}
	\vspace{-2mm}
	\caption{The gradual drift interval w.r.t. active cases over time}
	\label{fig:gradinsuranceActivcases}
\end{figure}

We also presented these results to our contact at the insurance company. The analyst confirmed that during the period identified by the gradual drift interval, the insurance company trialled a ``preferred repairer'' policy for this particular insurance brand, which is a low-budget one. Essentially claimants were required to get their motor vehicle damage fixed by the insurance company's preferred repairer, instead of choosing their own repairer. This change was implemented progressively, starting with selected claims only. The rationale was to increase the process efficiency for the insurance company. This change proved to be effective, since the number of active cases dropped significantly after the drift (see Figure \ref{fig:gradinsuranceActivcases}). 
This is also confirmed by the average process cycle time, which decreased from 91 to 26 days after the drift. 

The position of the gradual drift detected by the method in \cite{Martjushev2015} did not correspond to any business process change, according to the business analyst.

%
%
%
%
%

\vspace{-\baselineskip}
\section{Threats to validity}
\label{sec:validity}


A potential threat to the validity of the evaluation is the reliance of the proposed methods for sudden and gradual drift detection on a window size. The latter can affect the obtained accuracy in terms of F-score and detection delay. This limitation is to a large extent addressed by the fact that the window size is automatically adapted as the method progresses through the stream of traces. Thus, any effects of the initial choice of window size are evened out. We have also tested with different initial window sizes and provided guidelines for initializing this parameter.


In order to filter out sporadic and non-statistically significant changes in process behavior, we require that the P-value of at least $\phi$ consecutive statistical tests lie under the threshold. The $\phi$ parameter has been set empirically by striking the best trade-off of accuracy and detection delay on a subset of the datasets used for evaluation. We also tested an alternative technique -- Simple Moving Average (SMA) smoothing -- and obtained similar results, indicating that the reliance on $\phi$ is not a critical threat to validity.

Another possible threat to validity is that we evaluated our method over only two real-life logs and the results over these logs were validated by only one domain expert. To overcome this issue, we carried out an extensive evaluation over synthetic datasets. These datasets were generated using a systematic method aimed at testing the proposed algorithms on the 12 simple change patterns identified in \cite{Weber2008} and nested compositions thereof. Finally, the time performance measurements were averaged 9 times for each event log, covering several event log sizes and drift distances.



\vspace{-.5\baselineskip}
\section{Conclusion}\label{sec:conclusion}

This article outlined an automated method for detecting sudden and gradual drifts in business processes from execution traces.
An evaluation over synthetic logs showed that the method accurately discovers typical process changes and nested compositions thereof, outperforming a state-of-the art baselines both for sudden and gradual drifts. 
A separate evaluation on a large real-life log demonstrated the method's ability to detect drifts that correspond to user-recognizable process changes, as well as its scalability. 

The gradual drift detection method relies on the assumption that a gradual drift is delimited by two consecutive sudden drifts, such that the distribution of runs in-between these two drifts is a linear mixture of the distributions of runs before the first drift and after the second drift.
The accuracy achieved by the method in the experimental evaluation suggests that this assumption generally holds in practice. However, the assumption can be violated in some cases, for example when two gradual drifts overlap, in which case there might not be only two, but more sudden drifts observed during the two gradual drifts. The assumption is also likely to break if a sudden drift occurs in the middle of a gradual drift. Designing a more sophisticated method that would lift this assumption is an avenue for future work.

In its present form, the proposed method assumes that the input event log consists of a sequence of event labels, each representing the execution of one activity. Oftentimes, each event carries a payload containing data consumed or produced by the execution of the activity, or data about the resource who performed the activity. These payloads may help to better detect and characterize a business process drift. An avenue for future work is to enhance the proposed method by incorporating event payloads. The challenge is to encode the data payloads in a way that enables reliable statistical testing with a small number of runs per window. 

Another avenue for future research is to enhance our method by providing feedback that would help users understand the process change underpinning a detected drift. A possible direction to tackle this problem is to adapt our previous work on drift characterization \cite{ostovar2017characterizing}, which is based on measuring variations in the frequency of the behavioral relations observed in the log before and after the drift. A challenge in this context will be the characterization of change in the context of gradual drifts.



\section*{Acknowledgments}
\addcontentsline{toc}{section}{Acknowledgment} 
This research is  funded by the Australian Research Council Discovery Project DP150103356 and the Estonian Research Council.

\bibliographystyle{IEEEtran}
\bibliography{IEEEabrv,bibliography}  

%
\vspace{-3\baselineskip}
\begin{IEEEbiography}
  [{\includegraphics[width=25mm,height=32mm,clip,keepaspectratio]{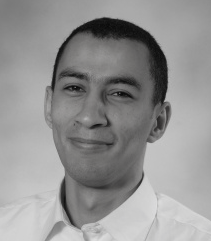}}]
  {Abderrahmane Maaradji} is a research fellow at the Queensland University of Technology. He obtained his PhD in computer science in 2011 at Universite Pierre and Marie Curie and Bell Laboratories, France. His research interests are in the area of process mining, with a focus on concept drift.
\end{IEEEbiography}
\vspace{-4\baselineskip}
\begin{IEEEbiography}
  [{\includegraphics[width=25mm,height=32mm,clip,keepaspectratio]{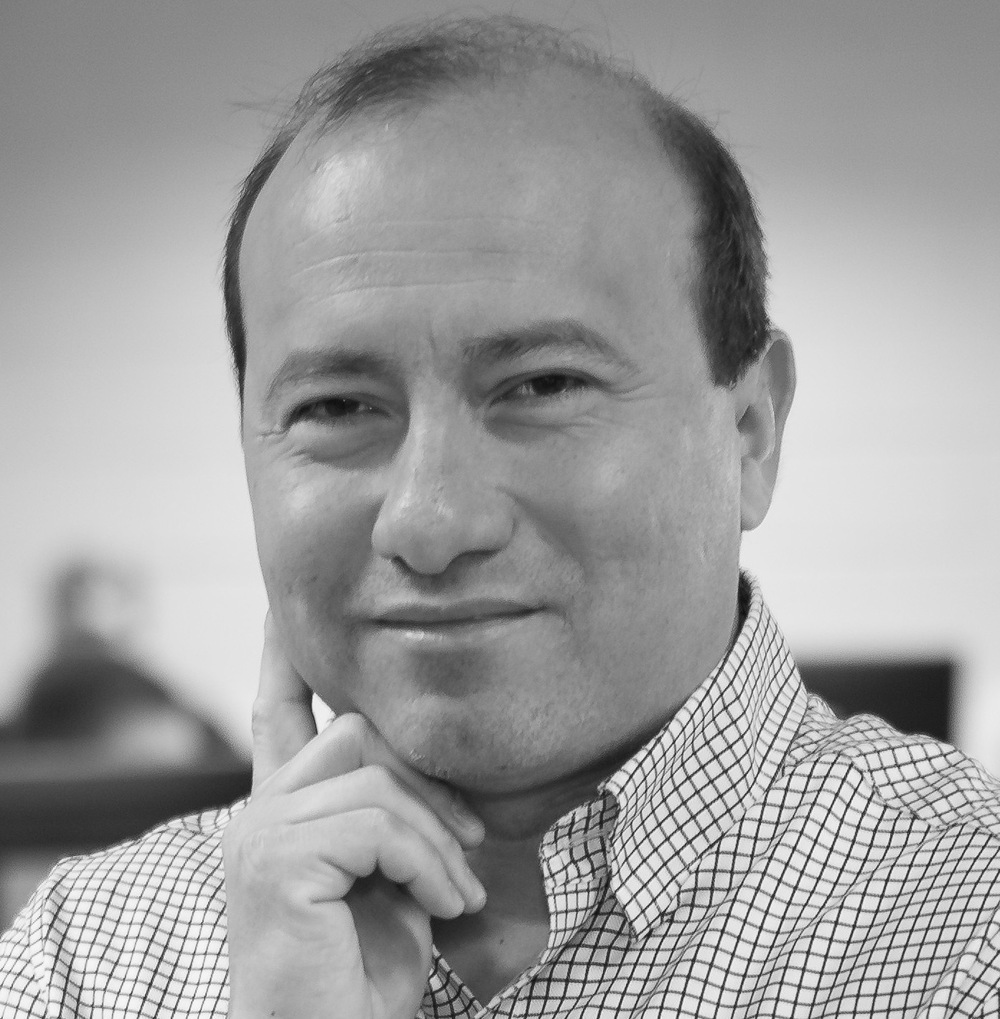}}]
  {Marlon Dumas} is Professor of Software Engineering and University of Tartu, Estonia and Adjunct Professor of Information Systems at Queensland University of Technology, Australia. His research focuses on combining data mining and formal methods for analysis and monitoring of business processes. He has published extensively across the fields of software engineering and information systems and has co-authored two textbooks on business process management.
\end{IEEEbiography}
\vspace{-3\baselineskip}
\begin{IEEEbiography}
  [{\includegraphics[width=25mm,height=32mm,clip,keepaspectratio]{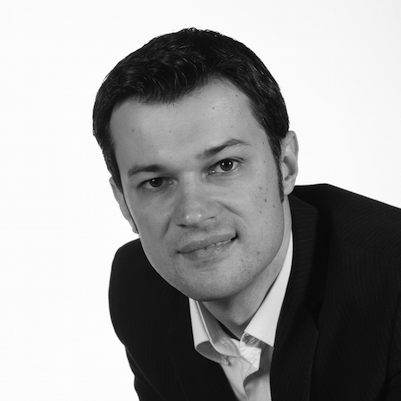}}]
  {Marcello La Rosa} is Professor of Information Systems at the Information Systems school of the Queensland University of Technology, Brisbane, Australia. His research interests include process consolidation, mining and automation, in which he published over 100 papers. He is co-author of the textbook Fundamentals of Business Process Management (Springer, 2013).
\end{IEEEbiography}
\vspace{-4\baselineskip}
\begin{IEEEbiography}
  [{\includegraphics[width=25mm,height=32mm,clip,keepaspectratio]{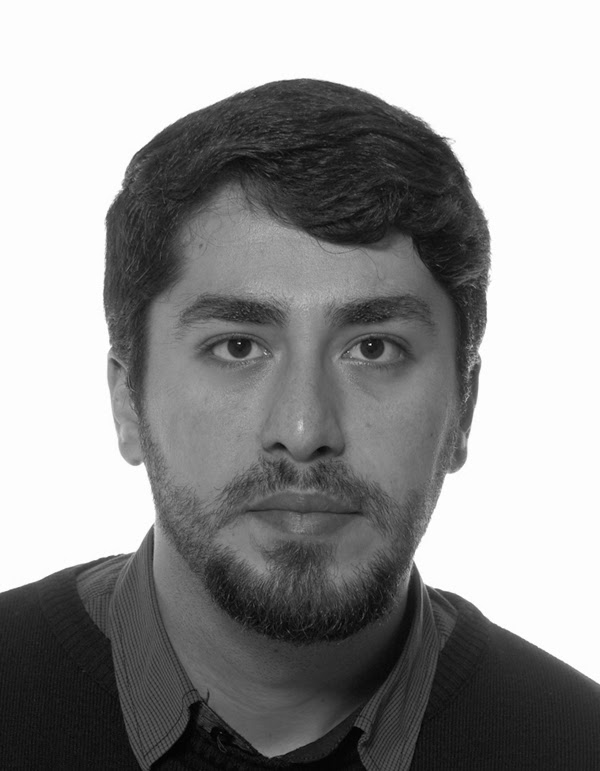}}]
  {Alireza Ostovar} is a PhD student at the Queensland University of Technology. His PhD topic spans the fields of business process management and process mining, with a focus on process drift detection and characterization.
\end{IEEEbiography}

\clearpage
\appendix[]
\label{appendix}

This Appendix provides a detailed description of both sudden and gradual drift detection algorithms and discusses their time complexity analysis. It also provides the separate plots of the recall, precision and F-score obtained for Figure \ref{fig:winsizeFscore} (Section \ref{sec:Fwinsize}).

\subsection*{Sudden drift detection algorithm}
\label{sec:suddenalgo}

Algorithm \ref{alg:sdrift} formally captures the sudden drift detection method. The algorithm has 
two parameters: (i) $\mathit{L}$: a stream of completed traces produced by the process; (ii) $\mathit{initW}$: initial size of the detection and reference windows; and (iii) $\mathit{maxBS}$: maximum available memory for the trace buffer storing the incoming traces, namely $\mathit{traceBuf}$. 

\SetAlFnt{\small}
\SetCommentSty{mycommfont}
\begin{algorithm}[hbt!]
\SetKwInput{Input}{Input}
\SetKwInOut{Output}{Output}
\DontPrintSemicolon
\Input{$L$: trace stream; $\mathit{initW}$: initial window size; $\mathit{maxBS}$: maximum buffer size.}

$\mathit{traceBuf, refTrace, detTrace}$ \tcp*[r]{Trace buffer, lists of traces within detection and reference windows, respectively}
$w \longleftarrow \mathit{initW}$ \tcp*[r]{Current window size}
$\mathit{detRuns}$, $\mathit{refRuns}$ \tcp*[r]{Lists of runs within detection and reference windows, respectively}
$A\_det$, $A\_ref$ \tcp*[r]{Sets of Alpha-relationships within detection and reference windows, respectively}  

$\mathit{chiThresh} \longleftarrow 0.05$ \tcp*[r]{Typical threshold value of chi-square test}
$\mathit{dTrace} \longleftarrow \mathit{NIL}$ \tcp*[r]{Current trace when P-value drops below $\mathit{chiThresh}$}
$\mathit{dW} \longleftarrow -1$ \tcp*[r]{Value of $w$ when P-value drops below $\mathit{chiThresh}$}
$\mathit{dLen} \longleftarrow 0$ \tcp*[r]{Number of consecutive tests that P-value remains below $\mathit{chiThresh}$}
$\phi \longleftarrow 1$ \tcp*[r]{Noise filter size when P-value drops below $\mathit{chiThresh}$}
\BlankLine

\While(\tcp*[f]{While it is possible to read new trace from the stream}){canRead(L)}{
	$\sigma \longleftarrow read(L)$\tcp*[r]{Read a new trace $\sigma$}
	\If{$size(\mathit{traceBuf}) = \mathit{maxBS}$}{
		$shift(\mathit{traceBuf})$\;
	}
	$insert(\mathit{traceBuf}, \sigma)$\;
	\If{$\text{length} (\mathit{traceBuf}) \geq 2 \cdot w$}{
		$\mathit{(refTrace,detTrace)} \longleftarrow extract(\mathit{traceBuf},w)$\;
		$(\mathit{A\_ref}, \mathit{A\_det}) \longleftarrow updateAlphaRelations(\mathit{refTrace, detTrace}, \mathit{A\_ref}, \mathit{A\_det})$\;
		$(\mathit{newRefRuns}, \mathit{newDetRuns}) \longleftarrow updateRuns(\mathit{refTrace, detTrace}, \mathit{refRuns}, \mathit{detRuns}, \mathit{A\_ref}, \mathit{A\_det})$\;
		$\mathit{w} \longleftarrow adWin(\mathit{newRefRuns}, \mathit{newDetRuns}, \mathit{refRuns}, \mathit{detRuns}, w)$\;
		$(\mathit{refRuns}, \mathit{detRuns}) \longleftarrow ( \mathit{newRefRuns}, \mathit{newDetRuns})$\;
		$\mathit{cMat} \longleftarrow contingencyMatrix(\mathit{refRuns}, \mathit{detRuns})$\;
		$\mathit{pVal} \longleftarrow chiSquareTest(\mathit{cMat})$\;
		\If{$\mathit{pVal} < \mathit{chiThresh}$}{
			$\mathit{dLen} \longleftarrow \mathit{dLen} + 1$\;
			\If{$\mathit{dTrace} = \mathit{NIL}$}{
				$\mathit{dTrace} \longleftarrow \sigma$\;
				$\mathit{dW} \longleftarrow w$\;
				$\phi \longleftarrow \mathit{dW}/3$\;
			}			
			\If{$\mathit{dLen} = \phi$}{
				$reportDrift(\mathit{dTrace})$ \tcp*[r]{Drift detected and reported}
			}
		}\Else{
			$\mathit{dTrace} \longleftarrow \mathit{NIL}$\;
			$\mathit{dW} \longleftarrow -1$\;
			$\mathit{dLen} \longleftarrow 0$\;
			$\Phi \longleftarrow 1$\;
		}
	}
}
\caption{Detect Sudden Concept Drifts}\label{alg:sdrift}
\end{algorithm}

Thus, each time a new trace $\sigma$ arrives, we first check if the buffer has reached its maximum size, and if so we shift the trace in the buffer and discard the least recent trace (lines 11-13). We then insert the new trace into the buffer (line 14). Only when the number of traces in the buffer reaches $2 \times \mathit{w}$ then the first statistical test can be performed (line 15). The reference and detection windows are then extracted from the $\mathit{traceBuf}$ based on the windows size $\mathit{w}$ (line 16).
The $\alpha$ relations are extracted into the matrices $\mathit{A\_ref}$ and $\mathit{A\_det}$ for the first time from the reference and detection windows $(refTrace,detTrace)$ respectively, and then incrementally updated (line 17). These relations are then used to update the runs and store them in temporary variables $\mathit{newRefRuns}$ and $\mathit{newDetRuns}$ (line 18). Next, as detailed in Section~\ref{sec:adwin}, we adjust the window size based on the evolution ratio $evolutionRatio$ (line 19). 
We then perform the Chi-square test of independence on the contingency matrix obtained from $\mathit{refRuns}$ and $\mathit{detRuns}$ (lines 20-22). If the resulting P-value of the Chi square test is below the threshold $\mathit{chiThresh}$ then a drift is asserted provided that the P-value remains under the threshold for $\phi$ consecutive statistical tests (lines 23-30). This latter condition is intended to avoid reporting incidental drops in P-value (oscillations) as discussed in Section~\ref{sec:oscillation}.



\textbf{Time complexity analysis:}
As explained in previous section, each time we handle a new trace from the stream, the algorithm performs three steps: \emph{first} it detects the concurrency relations in each sliding window; \emph{second} it updates the set of runs; and \emph{third} it performs the chi-square test. The complexity of the first step depends on the applied concurrency oracle. In our case, we use the $\alpha$ concurrency oracle that can be incrementally updated. Hence, the $\alpha$ relations of each one of the two first sliding windows (cf.\ $\mathit{A\_ref}$ and $\mathit{A\_det}$ in the algorithm) is computed. This operation only needs to be done once when the drift detection procedure is initialized. Subsequently, for each new trace read from the stream, or dropped from the sliding windows, each matrix is incrementally updated with direct access. Hence, the complexity of the \emph{first} step is then constant. In the second step, for each partially ordered run of the $w$ runs in the reference then detection windows, the algorithm iterates over the partial order and updates it based with direct access to the corresponding $\alpha$ relations matrix. Consequently, in the worst case, the time complexity of the \emph{second} step is $O(w * max\_trace\_length)$, where $max\_trace\_length$ is the length of the longest trace that bound the length of the longest run.
Regarding the third step, the complexity of the chi-square statistical test is linear to the number of columns of the contingency matrix which in the worst case is $w$. Thus, the complexity of the \emph{third} step is $O(w)$.
To sum up, for each new trace in the stream, the complexity of Algorithm~\ref{alg:sdrift} is the maximum complexity of its three steps, i.e. $O(w * max\_trace\_length)$. 

\subsection*{Gradual drift detection algorithm}
\label{sec:gradualalgo}

Algorithm \ref{alg:gdrift} formally captures the gradual drift detection method. The algorithm has two parameters: $SList$: the list of detected sudden drifts (i.e. their location in terms of time stamp or trace index in the stream), $HList$: The list of histograms. When  $SList$ can be directly obtained from the reported drifts of Algorithm~\ref{alg:sdrift}, the $HList$ can be reported by Algorithm~\ref{alg:sdrift} with a minor modification. A histogram can be built by incremental updates when a new trace in the stream is transformed into a partially ordered run at line 19 of Algorithm~\ref{alg:sdrift}.
It is worth mentioning that both sudden and gradual drift detection algorithms can run in parallel and communicate through a queue that stores no more than two sudden drifts and the corresponding three histograms.


For every pair of consecutive sudden drifts, Algorithm~\ref{alg:gdrift} reads the histograms before the first sudden drift, in-between the two drifts, and the one after the second sudden drift and up to the next drift or the end of the log (line 3-5). Then, it calculates the degree of freedom $df$ which equals the number of distinct runs (size of any of the histogram) minus one (line 6). Next, given $df$ and ${Tr}_{Chi}= 0.05 $, it uses the Chi-square table to fetch the critical value $chiCriticalValue$ (line 7). Consequently, all required inputs are ready to construct the inequality \ref{inequality} (line 8) and pass it to the solver (line 9). If the solver returns at least one solution $\left( x_0, y_0\right)$, then we declare a gradual drift between the two sudden drifts. 

\SetAlFnt{\small}
\SetCommentSty{mycommfont}
\begin{algorithm}[hbt!]
\SetKwInput{Input}{Input}
\SetKwInOut{Output}{Output}
\DontPrintSemicolon
\Input{$SList$: Sudden drifts list; $HList$ Histograms list}


${Tr}_{Chi} \longleftarrow 0.05$ \tcp*[r]{Typical threshold value of Chi-square test}
\BlankLine

\For{$i = 0, .., size(SList)-2 $}{
	$beforeHis \longleftarrow HList(i)$\tcp*[r]{read the histogram before the candidate gradual drift}
	$inHis \longleftarrow HList(i+1)$\tcp*[r]{read the histogram of the candidate gradual drift}
	$afterHis \longleftarrow HList(i+2)$\tcp*[r]{read the histogram after the candidate gradual drift}	
	$df \longleftarrow size(beforeHis)-1$\tcp*[r]{the degree of freedom is the size of any of the histograms}	
	$chiCriticalValue \longleftarrow chiSquareTable({Tr}_{Chi}, df)$\;
	$ineq \longleftarrow inequality(beforeHis, inHis, afterHis, chiCriticalValue)$\;
	$\left( x_0, y_0\right) \longleftarrow solver(ineq)$\;
	\If{$\left( x_0, y_0\right)$ not null}{
				$reportDrift(SList(i),SList(i+1))$ \tcp*[r]{Drift detected and reported}
	}
}
\caption{Detect Gradual Concept Drifts}\label{alg:gdrift}
\end{algorithm}

\textbf{Time complexity analysis: }
In order to detect gradual drifts, we first need to detect the sudden drifts and then assess the likelihood of each two consecutive sudden drifts being the start and the end of a gradual drift. 
The time complexity of the sudden drift detection method over a stream of traces has already been presented in Section \ref{sec:suddenalgo}. In the following we will only discuss the complexity of post-processing step for gradual drift detection. In this step three histograms are maintained with the frequencies of distinct runs before, within, and after the potential gradual drift. These histograms are direct-access data structures, and thus take constant time. These three histograms are then used to construct a two-variable inequality as an input to a solver. The time complexity of solving the inequality is determined by the complexity of the two-variable inequality solver employed, which in any case is independent of the size of the event log and of the size of the event label set.

\newpage

\subsection*{Impact of window size on the recall, precision and F-score}

\begin{figure}[htb!]
	\centering 
	\includegraphics[width=\linewidth]{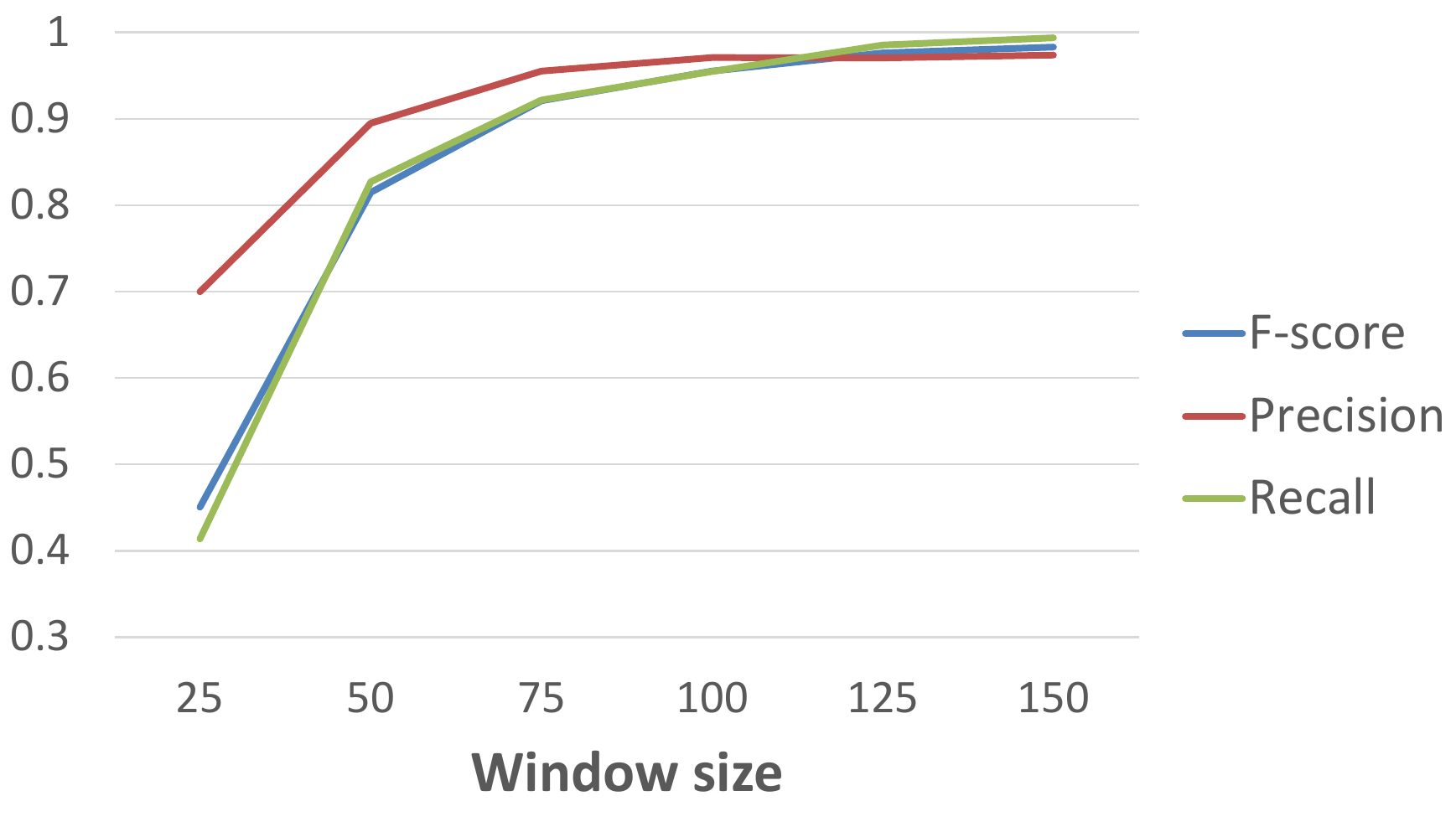}
	\vspace{-2mm}
	\caption{Recall, precision and F-score obtained with different fixed window sizes, averaged over all 72 event logs described in Section~\ref{sec:patterns}.}\label{fig:fscoreF_PR}
	\vspace{-\baselineskip}
\end{figure}

In Section \ref{sec:Fwinsize} we reported the accuracy of our sudden drift detection method in terms of the F-score obtained when we vary the size of the reference/detection window. While in Section \ref{sec:Fwinsize} we broke down the results based on the different event logs sizes in order to focus on the effect of window size on F-Score, in Figure \ref{fig:fscoreF_PR} we break down the F-Score into recall and precision to highlight the effect of window size on each specific measure (F-Score, recall and precision). In this figure, the value of each measure at a given window size is averaged over the 72 event logs used in our evaluation for sudden drift detection (cf. Section~\ref{sec:patterns}), with a fixed window size ranging from 25 to 150 traces. From this figure we observe that our sudden drift detection method errs similarly in both false positives (precision) and false negatives (recall). Unsurprisingly, as mentioned in Section \ref{sec:Fwinsize}, for a small window size (25 traces), the statistical test hardly converges. This results in a higher rate of missed drifts (lower recall).

\end{document}